\documentclass{new_tlp}
\usepackage{mathptmx} % standard

\usepackage{amssymb,latexsym,amsmath} % hp

\usepackage{times} % hp
\usepackage{graphicx} % hp
\usepackage{color}
\usepackage{xspace}

\usepackage{calrsfs}
\DeclareMathAlphabet{\pazocal}{OMS}{zplm}{m}{n}

\newtheorem{theorem}{Theorem}
\newtheorem{propositioni}[theorem]{Proposition}
\newenvironment{proposition}
   {\begin{propositioni}\rm}{\end{propositioni}}

\newtheorem{lemmai}[theorem]{Lemma}
\newenvironment{lemma}
   {\begin{lemmai}\rm}{\end{lemmai}}

\newtheorem{propertyi}[theorem]{Property}

\newtheorem{postulatei}[theorem]{Postulate}

\newtheorem{definitioni}{Definition}
\newenvironment{definition}
   {\begin{definitioni}\rm}{\end{definitioni}}

\newtheorem{corollaryi}[theorem]{Corollary}

\newtheorem{defii}[theorem]{Definition}

\newtheorem{claimi}[theorem]{Claim}

\newtheorem{conjecturei}[theorem]{Conjecture}

\newtheorem{exercisei}[theorem]{Exercise}

\newtheorem{questioni}[theorem]{Open Question}

\newtheorem{conventioni}[theorem]{Convention}

\newtheorem{facti}[theorem]{Fact}

\newtheorem{problemi}[theorem]{Problem}

\newtheorem{examplei}[theorem]{Example}

\newtheorem{remarki}[theorem]{Remark}
\newenvironment{example}
   {\begin{examplei}\rm}{\end{examplei}}

\newtheorem{notationi}[theorem]{Notation}

\newtheorem{claim2}{Claim}

\newcounter{tbsnr}
\newenvironment{tbs}
{\addtocounter{tbsnr}{1}\par\bigskip \noindent\fbox{\thetbsnr}
\hspace*{\fill}\begin{minipage}{0.8\textwidth}\tt}
{\end{minipage}\hspace*{\fill}\bigskip}

\definecolor{newgrs}{RGB}{0,0,0}%{0,129,129}
\definecolor{newpost}{RGB}{0,0,0}%{255,0,0}

\newcommand{\DLP}{\emph{DeLP}\xspace}
\newcommand{\DLPgr}{\emph{DeLP$^{\mathsf{(GR)}}$}\xspace}
\newcommand{\delp}{\DLP-program}

\newcommand{\no}{\ensuremath{\sim\!\!}\xspace}

\newcommand{\saltear}[1]{}

\newcommand{\dlp}{\textit{dlp}\xspace}

\newcommand{\comentario}[1]{}
\newcommand{\skippar}[1]{}

\newlength{\EA}  % Espacio Adicional
\newlength{\MEA} % Medio Espacio Adicional

\newtheorem{Notax}{{\bf Observaci\'on}}[section]

\newtheorem{obsx}{{\bf Observaci\'on}}[section]

\newtheorem{Algox}{{\sc Algorithm}}

\newcommand{\defleft}{\mbox{$\,-\hspace{-8pt}- \hspace{-4pt}\prec$ }}
\newcommand{\defleftarrow}{ {\ensuremath{\footnotesize \defleft}} }

\newcommand{\facto}[1]{\ensuremath{\mathit{#1}}}
\newcommand{\srule}[2]{\ensuremath{\mathit{#1}\gets\mathit{#2}}\xspace}
\newcommand{\drule}[2]{\ensuremath{\mathit{#1}\;\defleftarrow\mathit{#2}}\xspace}

\newcommand{\seq}[1]{\mbox{$\lfloor #1\rfloor$}}

\newcommand{\argnotation}[1]{\pazocal{#1}}

\newcommand{\Argum}[1]{\ensuremath{#1}\xspace}

\newcommand{\Arg}{\ensuremath{\Argum{\argnotation{A}}}\xspace}
\newcommand{\BArg}{\mbox{$\Argum{\argnotation{B}}$}\xspace}

\newcommand{\ie}{\textnormal{\emph{i.e.}, }}

\newcommand{\PP}{\ensuremath{\mathcal P}}
\newcommand{\DD}{\ensuremath{\Delta}}
\newcommand{\SSet}{\ensuremath{\Pi}}
\newcommand{\SD}{\mbox{$(\SSet,\DD)$}}

\newcommand{\pair}[2]{$(#1,#2)$}
\newcommand{\SyA}{\ensuremath{\SSet \cup \Arg}}

\newcommand{\AL}{\ensuremath{\langle \Arg, L \rangle}}

\newcommand{\BQ}{\mbox{$\langle \Barg,Q \rangle $}}
\newcommand{\CP}{\mbox{$\langle \Carg,P \rangle $}}
\newcommand{\Tree}[1]{\mbox{${\mathcal{T}}_{\small #1}$}}

\newcommand{\MTree}[1]{\mbox{\ensuremath{{\mathcal T^*}_{\hspace*{-4pt}\small #1}}}}
\newcommand{\Aline}{\mbox{$\Lambda$}}
\newcommand{\Sline}{\mbox{$\Aline_S$}}
\newcommand{\Iline}{\mbox{$\Aline_I$}}

\newcommand{\Ar}[2]{\mbox{$\langle #1,\mathit{#2} \rangle $}}

\newcommand{\La}{\mbox{$L_1$}}

\newcommand{\ALa}{\mbox{$\langle \Arg_1,L_1 \rangle $}}
\newcommand{\ALb}{\mbox{$\langle \Arg_2,L_2 \rangle $}}
\newcommand{\ALc}{\mbox{$\langle \Arg_3,L_3 \rangle $}}
\newcommand{\ALd}{\mbox{$\langle \Arg_4,L_4 \rangle $}}
\newcommand{\ALe}{\mbox{$\langle \Arg_5,L_5 \rangle $}}
\newcommand{\ALi}{\mbox{$\langle \Arg_i,L_i \rangle $}}
\newcommand{\ALk}{\mbox{$\langle \Arg_k,L_k \rangle $}}

\newcommand{\ALiAnt}{\mbox{$\langle \Arg_{i-1},L_{i-1} \rangle $}}
\newcommand{\ALiProx}{\mbox{$\langle \Arg_{i+1},L_{i+1} \rangle $}}
\newcommand{\ALn}{\mbox{$\langle \Arg_n,L_n \rangle $}}

\newcommand{\Aarg}{\mbox{$\Argum{\argnotation{A}}$}\xspace}
\newcommand{\Barg}{\mbox{$\Argum{\argnotation{B}}$}\xspace}
\newcommand{\Carg}{\mbox{$\Argum{\argnotation{C}}$}\xspace}

\newcommand{\BQa}{\mbox{$\langle \Barg_1,Q_1 \rangle $}}
\newcommand{\BQb}{\mbox{$\langle \Barg_2,Q_2 \rangle $}}
\newcommand{\BQi}{\mbox{$\langle \Barg_i,Q_i \rangle $}}
\newcommand{\BQk}{\mbox{$\langle \Barg_k,Q_k \rangle $}}
\newcommand{\Dnode}{\mbox{``$D$''}}
\newcommand{\Unode}{\mbox{``$U$''}}

\long\def\symbolfootnote[#1]#2{\begingroup%
	\def\thefootnote{\fnsymbol{footnote}}\footnote[#1]{#2}\endgroup}
\long\def\symbolfootnote[#1]#2{\begingroup%
	\def\thefootnote{\fnsymbol{footnote}}\footnote[#1]{#2}\endgroup}

\newcommand{\Arga}{\mbox{$\Aarg_1$}}
\newcommand{\Argb}{\mbox{$\Aarg_2$}}
\newcommand{\Argc}{\mbox{$\Aarg_3$}}
\newcommand{\Argd}{\mbox{$\Aarg_4$}}

\newcommand{\Argi}{\mbox{$\Aarg_i$}}

\usepackage{xspace}

\newcommand\eg{\emph{e.g.,}\xspace}

\newcommand\bcmdtab{\noindent\bgroup\tabcolsep=0pt%
  \begin{tabular}{@{}p{10pc}@{}p{20pc}@{}}}
\newcommand\ecmdtab{\end{tabular}\egroup}

  \title[Theory and Practice of Logic Programming]
	{A Comparative Study of Some Central Notions of \emph{ASPIC}$^+$ and \DLP}

 \author[Alejandro J. Garc\'{\i}a, Henry Prakken and Guillermo R. Simari]
         {ALEJANDRO J. GARC\'IA\\
            Department of Computer Science \& Engineering\\
            Universidad Nacional del Sur, Bahia Blanca, Argentina\\
                      \email{ajg@cs.uns.edu.ar}
          \and HENRY PRAKKEN\\
            Department of Information and Computing Sciences, Utrecht University \& \\
Faculty of Law, University of Groningen \\
The Netherlands\\
\email{h.prakken@uu.nl}
         \and GUILLERMO R. SIMARI\\
            Department of Computer Science \& Engineering\\
            Universidad Nacional del Sur, Bahia Blanca, Argentina\\
            \email{grs@cs.uns.edu.ar}
}

\jdate{December 2018}
\pubyear{2019}
\newcommand{\begintab}[1]{ \begin{tabular}{#1} }
\newcommand{\btab}[1]{\medskip \par \begintab{#1}}
\newcommand{\etab} {\end{tabular} \medskip \par \noindent }

\newcommand{\ben}{\begin{enumerate}}
\newcommand{\een}{\end{enumerate}}
\newcommand{\bit}{\begin{itemize}}
\newcommand{\eit}{\end{itemize}}

\newcommand{\nm}{\mid\joinrel\sim}

\newcommand{\imp}{\rightarrow}
\newcommand{\Imp}{\Rightarrow}

\newcommand{\K}{\pazocal{K}}
\newcommand{\A}{\pazocal{A}}

\newcommand{\C}{\pazocal{C}}
\newcommand{\R}{\pazocal{R}}

\newcommand{\LA}{\pazocal{L}}

\newcommand{\ASPIC}{\emph{ASPIC}$^+$}

\begin{document}

\label{firstpage}

\maketitle

  \begin{abstract}
This paper formally compares some central notions from two well-known formalisms for rule-based argumentation, \DLP and \ASPIC. The comparisons especially focus on intuitive adequacy and inter-transla\-tability, consistency, and closure properties.
As for  differences in the definitions of arguments and attack, it turns out that \DLP's definitions are intuitively appealing but that they may not fully comply with Caminada and Amgoud’s rationality postulates of strict closure and indirect consistency.
For some special cases, the \DLP definitions are shown to fare better than \ASPIC.
Next, it is argued that there are reasons to consider a variant of \DLP with grounded semantics, since in some examples its current notion of warrant arguably has counterintuitive consequences and may lead to sets of warranted arguments that are not admissible.
Finally, under some minimality and consistency assumptions on \ASPIC\ arguments,
a one-to-many correspondence between \ASPIC arguments and \DLP arguments is identified in such a way that if the \DLP warranting procedure is changed to grounded semantics, then \DLP's notion of warrant and \ASPIC's notion of justification are equivalent.
This result is proven for three alternative definitions of attack.
  \end{abstract}


\tableofcontents

%%%%%%%%%%%%%%%%%%%%%%%%%%%%%%%%%%%%%%%%%%%%%%%%%%%%%%%
\section{Introduction}

\ASPIC~\cite{hp10aspicJAC} and Defeasible Logic Programming, or \DLP for short~\cite{g+s04}, are two well-known rule-based formalisms for argumentation. `Rule-based' is not about expressiveness but about how arguments are constructed. In a rule-based approach\footnote{This and some other fragments in this paper are taken (or adapted) from \citeN{m+p18}.}, arguments are formed by chaining applications of inference rules into inference trees or graphs. This approach can be contrasted with approaches defined in terms of logical consequence notions,  in which arguments are premises-conclusion pairs where the premises are consistent and imply the conclusion according to the consequence notion of some adopted `base logic'. Examples of this approach are classical-logic argumentation \cite{cay95,b+h01,b+h08,g+h11} and its generalisation into abstract Tarskian-logic argumentation \cite{a+b13}. It is important to note that, unlike these logic-based approaches, rule-based approaches in general do not adopt a single base logic but two base logics, one for the strict and one for the defeasible rules.

\ASPIC\ and \DLP are similar in various respects: both have a distinction between strict and defeasible inference rules and both use preferences to resolve attacks into defeats.  The shared rule-based approach and these further similarities warrant a detailed comparison between the two frameworks. Such a comparison is the topic of this paper. It will turn out that there are also differences, the main one being that while \ASPIC\ evaluates arguments with the by now standard Dungean semantics of abstract argumentation frameworks \cite{dung95}, \DLP has a special-purpose definition of argument evaluation. Both of \ASPIC\ and of \DLP various versions exist. As for \DLP, we will discuss the version introduced by \citeN{g+s04}, which arguably is the standard version. As for \ASPIC\ we will unless indicated otherwise assume the version of \citeN{m+p13aij} with defeat-conflict-freeness, no consistency constraints on premise sets and the contrariness relation corresponding to `strong' or `symmetric' negation.
%This version is also described in Section~2.2 of \citeN{m+p18}.

We will compare \DLP 2004 with \ASPIC\ 2013, and we will also study modifications of both systems with ideas from the other systems.
In particular, we will consider a version of \DLP with grounded semantics and a version of \ASPIC\ with \DLP's notion of rebutting attack; either with or without \DLP's consistency constraints on arguments.
% grs ---- 	(By the way, with the current setup of the paper, changing the title to 	"A comparison between ASPIC+ and DeLP-with-grounded semantics does not
% grs ---- 	seem to be appropriate, since we do many other things.)
% grs ---- AGREED
Just before this paper was finished, we learned that \citeN{p+c18} had also carried out a comparison between \ASPIC and \DLP.   Nevertheless, our investigation can be regarded as complementary to theirs. In their work, they seek to revisit aspects that differentiate \DLP\ from \ASPIC, analyze the common ground between the two approaches, and study the possibility of establishing conditions that would help bridge the gap between them.  The discussion mainly centers on the similarities and differences between \ASPIC and \DLP regarding knowledge representation capabilities, the mechanism adopted for argument construction, and the different types of attack and defeat they consider. Their focus is not on formally proving properties of or relations between the two formalisms.

% -- grs To summarise our main findings, as for  differences in the definitions of arguments and attack, it will turn out that \DLP's definitions are intuitively appealing but that they may lead to a violation of the rationality postulates of strict closure and indirect consistency.
% -- grs For some special cases, the \DLP definitions will be shown to fare better than \ASPIC.
To summarise our main findings, as for  differences in the definitions of arguments and attack, it will turn out that \DLP's definitions are intuitively appealing and in some special cases the \DLP definitions will be shown to fare better than \ASPIC.
%We will also elaborate on the situation regarding the possible conflicts
On the other hand, the \DLP definitions may not fully comply with the rationality postulates of strict closure and indirect consistency introduced by \citeN{c+a07}.
 In Subsection~\ref{Rationality.Postulates}, we will include a thorough discussion about these issues.
As we will discuss in Section~\ref{argeval}, while the \DLP definition of warrant is similar to grounded semantics, there are also differences, caused by the fact that the constraints on the argument evaluation do not coincide with the constraints on games in the game-theoretic proof theory for grounded semantics.
For these reasons, we will introduce a special version of \DLP under grounded semantics, since in some examples its current notion of warrant arguably has counterintuitive consequences and may lead to sets of warranted arguments that are not admissible under Dung's definition.
Finally, under some minimality and consistency assumptions on \ASPIC\ arguments,
a one-to-many correspondence between \ASPIC arguments and \DLP arguments will be identified in such a way that if the \DLP warranting procedure is changed to grounded semantics, then \DLP's notion of warrant and \ASPIC's notion of justification are equivalent.
This result will be proven for three alternative definitions of attack.

This paper is organised as follows. We start with a brief sketch of the history of both frameworks in Section~\ref{history} and a summary of the formalisms in Section~\ref{prelim}.
%%%
%\marginpar{\henry{Order of sections changed in this para}}
%%%
We then compare the argument and attack definitions of the two formalisms in Sections~\ref{comparing} and~\ref{attacksec}.
We will argue that \DLP's definitions are interesting alternatives to  \ASPIC\ definitions which in some special cases represent possible improvements.
Then in Section~\ref{argeval} we compare the different ways in which \ASPIC\ and \DLP evaluate arguments.
We will argue that some differences reveal possible drawbacks of the \DLP semantics.
After observing that the motivation behind \DLP's semantics is similar to the intuitions underlying \citeANP{dung95}'s~\citeyear{dung95} grounded semantics,  we propose a version of \DLP with grounded semantics, arguing that all the examples given by \citeN{g+s04} as reasons for their special semantics are treated as they want by grounded semantics.
Finally, in Section~\ref{corrsec} we prove correspondence results with respect to arguments, attacks, defeats, and extensions
%%%
%\marginpar{HP: `as modified with grounded semantics' deleted}
%%%
between \ASPIC\ and \DLP. 

%%%%%%%%%%%%%%%%%%%%%%%%%%%%%%%%%%%%%%%%%%%%%%%%%%%%%%%
\section{History}\label{history}

\ASPIC\ originated from the European ASPIC project as an attempt to integrate and consolidate the then state-of-the art in formal argumentation (see \citeN{aspic06}). It was in particular inspired by the research of \citeN{pol87}, \citeN{pol95} and \citeN{vre97}.
A basic version without preferences or premise attack was used by \citeN{c+a07} as a vehicle for introducing and studying various so-called rationality postulates for argumentation. \citeN{hp10aspicJAC} introduced the first full version of \ASPIC, introducing premise attack and preferences. Since then the framework has been further developed and studied in several publications. For a detailed overview see Section 5 of \citeN{m+p18}.
In consequence, \ASPIC\ as it has been developed over the years is not a single framework but a family of frameworks varying on several elements.

\DLP was developed on the basis of \citeN{s+l92}, who presented a rule-based argumentation system with both strict and defeasible inference rules, with specificity as a means to resolve attacks and with an argument evaluation definition taken from \citeN{pol87}, which was later by \citeN{dung95} shown to be equivalent to his grounded semantics. Inspired by this work, \DLP was developed in a series of papers, culminating in \citeN{g+s04}, which is now regarded as the standard paper on \DLP. 
The idea of argument evaluation in terms of a dialectical tree, now typical for \DLP, was introduced by \citeN{gcs93} and \citeN{scg94}. 
The first paper establishing conditions on the construction of the branches of a dialectical tree (called an argumentation line) was \citeN{gsc98}; thus, this was the paper that gave up grounded semantics for \DLP.

%%%%%%%%%%%%%%%%%%%%%%%%%%%%%%%%%%%%%%%%%%%%%%%%%%%%%%%

\section{Formal Preliminaries}\label{prelim}
%\margin{check this}
In this section, we summarise the formal systems used throughout the paper.
%%%
%\marginpar{\henry{slighly changed since we do give examples}}
%%%
More details can be found in the papers already mentioned above and in~\citeN{bcg11} and~\citeN{bcg18} for abstract argumentation frameworks,~\citeN{m+p14} and \citeN{m+p18} for \ASPIC, and~\citeN{g+s14} and \citeN{g+s18} for \DLP.
It is relevant at this point to remark that the presentations of \ASPIC\ and \DLP contained in this paper heavily rely on earlier presentations of these systems, such as the ones cited.

\subsection{Abstract Argumentation Frameworks}\label{sec:afs}

An \emph{abstract argumentation framework} ($AF$) is a pair $\langle \A,\mathit{attack}\rangle$, where $\A$ is a set of arguments and $\mathit{attack}$ $\subseteq$ $\A$ $\times$ $\A$. The theory of $AFs$ \cite{dung95} identifies sets of arguments (called \emph{extensions}) which are internally coherent and defend themselves against attack. An argument $A \in \A$ is \emph{defended} by a set by $S \subseteq \A$ if for all $B \in \A$: if $B$ attacks $A$, then some $C \in S$ attacks $B$.  A set $S$ of arguments is \emph{conflict-free} if no argument in $S$ attacks an argument in $S$. Then, relative to a given $AF$,
\\[-13pt]
\bit
\item $E$ is \emph{admissible} if $E$ is conflict-free and defends all its members;
\item $E$ is a \emph{complete extension} if $E$ is admissible and $A \in E$ iff $A$ is defended by $E$;
\item $E$ is a \emph{preferred extension} if $E$ is a $\subseteq$-maximal admissible set;
\item $E$ is a \emph{stable extension} if $E$ is admissible and attacks all arguments outside it;
\item $E \subseteq \A$ is the \emph{grounded extension} if $E$ is the least fixpoint of operator $F$, where $F(S)$ returns all arguments defended by $S$.
\eit
 Finally, for $T \in \{$complete, preferred, grounded, stable$\}$, $X$ is \emph{sceptically} or \emph{credulously} justified under the $T$ semantics if $X$ belongs to all, respectively at least one, $T$ extension.

In \ASPIC\ the attack relation is renamed to \emph{defeat} to distinguish it from a more basic notion of conflict between arguments, which in \ASPIC\ is called attack. Moreover, the following terminology is used. Argument $A$ \emph{strictly defeats} argument $B$ if $A$ defeats $B$ and $B$ does not defeat $A$. Argument $A$ \emph{weakly defeats} argument $B$ if $A$ defeats $B$ and $B$ defeats $A$.

In the comparisons with \DLP we will use grounded semantics. In particular, we will use the following game-theoretic proof theory, which is sound and complete with respect to grounded semantics \cite{hp99,m+c09}.

\begin{definition}\label{dialogue}
An {\em argument game} for grounded semantics is a finite nonempty sequence of moves
$move_{i} = (Player_{i}, Arg_{i})$ $(i > 0)$, such that
\begin{enumerate}
   \item $Player_{i} = P$ iff $i$ is odd; and $Player_{i} = O$
   iff $i$ is even;
   \item If $Player_{i} = Player_{j} = P$ and $i \not= j$,
   then $Arg_{i} \not= Arg_{j}$;
   \item If $Player_{i} = P$, then $Arg_{i}$ strictly defeats
   $Arg_{i-1}$;
   \item If $Player_{i} = O$, then $Arg_{i}$ defeats
   $Arg_{i-1}$.
\end{enumerate}
\end{definition}
The first condition says that the proponent begins and then the players take turns, while the second condition prevents the proponent from repeating its attacks. The last two conditions form the heart of the definition: they state the burdens of proof for $P$ and $O$. The non-repetition rule and the condition that $P$ moves strict defeaters (as opposed to $O$ being allowed to move any defeater) are not needed for soundness and completeness but they make many otherwise infinite games finite.
\begin{definition}
A player {\em wins an argument game} iff the other player cannot
move. An argument $A$ is {\em provably justified} iff the proponent has a winning strategy in a game beginning with $A$.
\end{definition}
As is well-known, a strategy for the proponent can be displayed as a tree of games which only branches after the proponent's moves and which then contains all defeaters of this move. A strategy for the proponent is then winning if all games in the tree end with a move by the proponent. We will use these observations below in comparing the grounded argument game with \DLP's dialectical trees.

\subsection{\emph{ASPIC}$^+$}

We next  specify the present paper's instance of the  \ASPIC\ framework. It defines abstract argumentation systems as structures consisting of a logical language $\LA$ with symmetric negation and two sets $\R_s$ and $\R_d$ of strict and defeasible inference rules defined over $\LA$.  In the present paper $\LA$ is a language of propositional or predicate-logic literals, since this is also the language assumed by DeLP. Arguments are constructed from a knowledge base  (a subset of $\LA$) by combining inferences over $\LA$.
Formally:
	\begin{definition}\label{as}[Argumentation System]
		An \emph{argumentation system} (AS) is a pair $AS=(\LA,\R)$ where:\\[-13pt]
		\begin{itemize}
			\item $\LA$ is a logical language consisting of propositional or ground predicate-logic literals
			\item $\R = \R_s \cup \R_d$ is a set of strict ($\R_s$) and defeasible ($\R_d$) inference rules of the form \linebreak
$\{\varphi_1$, \ldots, $\varphi_n\}$ $\to$ $\varphi$ and  $\{\varphi_1$, \ldots,
$\varphi_n\}$ $\Rightarrow$ $\varphi$ respectively (where $\varphi_i,\varphi$ are meta-variables ranging over wff in $\LA$), such that $\R_s \cap \R_d = \emptyset$. Here $\varphi_1,\ldots,\varphi_n$ are called the \emph{antecedents} and $\varphi$ the \emph{consequent} of the rule.\footnote{Below the brackets around the antecedents will usually be omitted.}
%			\item $n$ is a partial function such that $n : \R_d \longrightarrow \LA$.
		\end{itemize}
	\end{definition}
%Informally, $n(r)$ is a well-formed formula (wff) in $\LA$ which says that the defeasible rule $r \in \R$ is applicable, so that an argument claiming $\neg n(r)$ attacks an inference step in the argument using $r$.
We write $\psi = -\varphi$ just in case $\psi = \neg \varphi$ or $\varphi = \neg \psi$. Note that $-$ is not part of the logical language $\LA$ but a metalinguistic function symbol to obtain more concise definitions. Also, for any rule $r$ the \emph{antecedents} and \emph{consequent} are denoted, respectively, with $\mathtt{ant}(r)$ and $\mathtt{cons}(r)$.
%
%The set $\R_s$ is said to be {\bf closed under transposition} if whenever $\varphi_1,\ldots,\varphi_n \rightarrow \psi$ $\in \R_s$,
%then for $i = 1 \ldots n$, $\varphi_1, \varphi_{i-1}, \psi', \varphi_{i+1},$ $\ldots,\varphi_n \rightarrow \varphi'_i$ $\in \R_s$ for all $\varphi'_i$ such that $\varphi'_i = -\varphi_i$ and all $\psi'$ such that $\psi' = -\psi$.

The set $\R_s$ is said to be \emph{closed under transposition} if whenever $S \imp \varphi \in \R_s$,
then  $S \setminus \{s_i\} \cup -\varphi  \imp -s_i \in \R_s$ for any $s_i \in S$. This notion is important since many consistency and closure results in the literature depend on the condition that $\R_s$ is closed under transposition.

\begin{definition}\label{consistency}[\textbf{Consistency}]
For any $S \subseteq \LA$, let the \emph{closure of $S$ under strict rules}, denoted $Cl_{R_s}(S)$, be the smallest set containing $S$ and the consequent of any strict rule in $\R_s$ whose antecedents are in $Cl_{R_s}(S)$. Then a  set $S$ $\subseteq$ $\LA$ is \emph{directly consistent} iff $\nexists$ $\psi$, $\varphi$ $\in$ $S$ such that $\psi = -\varphi$, and \emph{indirectly consistent} iff $Cl_{R_s}(S)$ is directly consistent.
\end{definition}
Note that the notion of indirect consistency is relative to a given set of strict rules.
%Below we will often leave this set implicit, assuming it to be obvious from the context.
%In general arguments are in \ASPIC\ constructed by applying inference rules to a knowledge base of formulas from $\LA$. However, as will become apparent later, for purposes of comparison with DeLP it is convenient to represent facts as strict inference rules with empty antecedents, as also done in e.g.\ \citeN{c+a07,dung16}. This is possible since the `facts' in DeLP are not attackable, i.e., they are assumed certain; as in DeLP, attackable (i.e., uncertain) facts can be represented as defeasible rules with empty antecedent. This also simplifies the formal inductive definition of an argument, since the condition declaring each item from the knowledge base to be an argument can be omitted.
%
\begin{definition}\label{kb}[\textbf{Knowledge Bases}]
A \emph{knowledge base} over an $AS = (\LA,\R)$ is a set $\K \subseteq \LA$.
\end{definition}
In this paper $\K$ corresponds to the `necessary premises' in other \ASPIC\ publications, which are intuitively certain and therefore not attackable. Since the `facts' in DeLP are also not attackable, we assume in this paper that the set of attackable or `ordinary' premises from other \ASPIC\ publications is empty. We will, as is also usually done in DeLP,  represent what intuitively are uncertain premises $\varphi$ as defeasible rules $\Rightarrow \varphi$.
%We  also assume that no element of $\K$ occurs in the consequent of any rule in $\R$.
In what follows, for a given argument, the function $\mathtt{Prem}$ returns all the formulas of $\K$ (called \emph{premises}) used to build the argument,
%$\mathtt{Prop}$ returns all the propositions used in that argument, {\bf *** is Prop used at all? ***}
$\mathtt{Conc}$ returns its conclusion, $\mathtt{Sub}$ returns all its sub-arguments, $\mathtt{LDR}$ returns the last defeasible rules used in the argument,  $\mathtt{Rules}$ and $\mathtt{DefRules}$ return, respectively, all rules and all defeasible rules of the argument and $\mathtt{TopRule}$ returns the last rule used in the argument.
An argument is now formally defined as follows.

\begin{definition}\label{arg}[\textbf{Arguments}]
A \emph{argument} $A$ on the basis of a knowledge base $\K$ in an argumentation system $AS$ is a structure obtainable by applying one or more of the following steps finitely many times:\\[-13pt]
\begin{enumerate}
    \item\label{arg1} $\varphi$ if $\varphi$ $\in$ $\K$ with:
                $\mathtt{Prem}(A)       = \{\varphi\}$;
                $\mathtt{Conc}(A)       = \varphi$;
    %            $\mathtt{Prop}(A)       = \{\varphi\}$, \\
                $\mathtt{Sub}(A)        = \{\varphi\}$;
  %              $\mathtt{PropSub}(A)$        = $\emptyset$;
   %             $\mathtt{ImmSub}(A)$        = $\emptyset$;
  %              $\mathtt{DefRules}(A)$  = $\emptyset$;
		$\mathtt{LDR}(A)$  = $\emptyset$;\\
               $\mathtt{Rules}(A)$  = $\emptyset$;
                            $\mathtt{DefRules}(A)$  = $\emptyset$;
                $\mathtt{TopRule}(A)$  = undefined.
\item\label{arg2} $[\{A_1, \ldots, A_n\} \rightarrow \psi]$\footnote{The square brackets make the presentation of examples more concise. They and the curly brackets will be omitted if there is no danger for confusion.} if $A_1, \ldots , A_n$ are  arguments such that
%$\psi \not\in \mathtt{Conc}(\{A_1, \ldots , A_n\})$ and
$\mathtt{Conc}(A_1), \ldots,\mathtt{Conc}(A_n)$ $\rightarrow \psi \in \R_s$ with:\\
            $\mathtt{Prem}(A)       = \mathtt{Prem}(A_1) \cup \ldots \cup \mathtt{Prem}(A_n)$;\\
                $\mathtt{Conc}(A)       = \psi$;\\
 %               $\mathtt{Prop}(A)       = \mathtt{Prop}(A_1) \cup \ldots \cup \mathtt{Prop}(A_n) \cup \{\psi\}$, \\
                $\mathtt{Sub}(A)        = \mathtt{Sub}(A_1) \cup \ldots \cup \mathtt{Sub}(A_n) \cup \{A\}$;\\
 %               $\mathtt{PropSub}(A)        = \mathtt{Sub}(A_1) \cup \ldots \cup \mathtt{Sub}(A_n)$,\\
   %             $\mathtt{ImmSub}(A)        = \{A_1, \ldots , A_n\}$,\\
                $\mathtt{LDR}(A)$ $=$ $\mathtt{LDR}(A_1) \cup \ldots \cup \mathtt{LDR}(A_n)$;
                $ \mathtt{Rules}(A)$ $=$ $\mathtt{Rules}(A_1) \cup \ldots \cup \mathtt{Rules}(A_n) \cup$\\
                $ \{\mathtt{Conc}(A_1), \ldots,\mathtt{Conc}(A_n) \rightarrow \psi\}$;\\
                             $\mathtt{DefRules}(A)$ $=$ $\mathtt{Rules}(A) \cap \R_d$;\\
   $\mathtt{TopRule}(A)$  = $\mathtt{Conc}(A_1), \ldots,\mathtt{Conc}(A_n) \rightarrow \psi$.	
\item\label{arg3} $[\{A_1, \ldots, A_n\} \Rightarrow \psi]$ if
$A_1, \ldots , A_n$ are  arguments such that
%$\psi \not\in \mathtt{Conc}(\{A_1, \ldots , A_n\})$ and
$\mathtt{Conc}(A_1), \ldots,\mathtt{Conc}(A_n)$ $\Rightarrow \psi \in \R_d$, with: \\
              $\mathtt{LDR}(A)$ $=$ $\{\mathtt{Conc}(A_1), \ldots,\mathtt{Conc}(A_n)$ $\Rightarrow \psi\}$;\\
								$ \mathtt{Rules}(A)$ $=$ $\mathtt{Rules}(A_1) \cup \ldots \cup \mathtt{Rules}(A_n) \cup \{\mathtt{Conc}(A_1), \ldots,\mathtt{Conc}(A_n)$ $\Rightarrow \psi\}$;\\
								$\mathtt{TopRule}(A)$  = $\mathtt{Conc}(A_1), \ldots,\mathtt{Conc}(A_n) \Rightarrow \psi$\\	
and the other notions defined as in (2).
           \end{enumerate}
An argument $A$ is \emph{strict} if  $\mathtt{DefRules}(A)$ $= \emptyset$, otherwise $A$ is \emph{defeasible}.
%And $A$ is \emph{firm} if  $\mathtt{Prem}(A) \subseteq \K_n$, otherwise $A$ is \emph{plausible}.
\end{definition}
	\noindent When $\mathtt{Conc}(A) = \varphi$ we sometimes say that $A$ is an argument for $\varphi$. Each of the functions $\mathtt{Func}$ in this definition is also defined on sets of arguments $S=\{A_1,\ldots,A_n\}$ as follows: $\mathtt{Func}(S)=\mathtt{Func}(A_1)\cup\ldots\cup\mathtt{Func}(A_n)$. Note that we overload the $\rightarrow$ and $\Rightarrow$ symbols to denote an argument while they also denote strict, respectively, defeasible inference rules. This is common practice in argumentation and originates from \citeN{vre97}.

\begin{example}\label{exargument}
Consider a knowledge base in an argumentation system with $\LA$ consisting of $p,q,r,s,t,u,v,x$ and their negations, with $\R_s = \{s_1,s_2\}$ and $\R_d = \{d_1,d_2,d_3\}$, where:
\btab{llllll}
$d_1$: & $p \Imp q$ &&&     $s_1$: & $p,q \imp r$\\
$d_2$: & $\Imp t$ &&&       $s_2$: & $t \imp \neg q$      \\
$d_3$: & $v,x \Imp \neg t$   &&&  $s_3$: & $u \imp v$
\etab
Let $\K = \{p,u,x\}$. An argument $A_3$ for $r$ (i.e., with conclusion $r$) with subarguments $A_1$ for $p$ and $A_2$ for $q$   is displayed in Figure~\ref{ex:argument}, with the premises at the bottom and the conclusion at the top of the tree. In this and the next figure, strict inferences are indicated with solid lines while defeasible inferences and rebuttable conclusions are displayed with dotted lines. The figure also displays the formal structure of the argument. Note that the argument can also be written as $[p, [p \Imp q] \imp r]$.
\begin{figure}[ht]
\centering
				\includegraphics[scale=0.60]{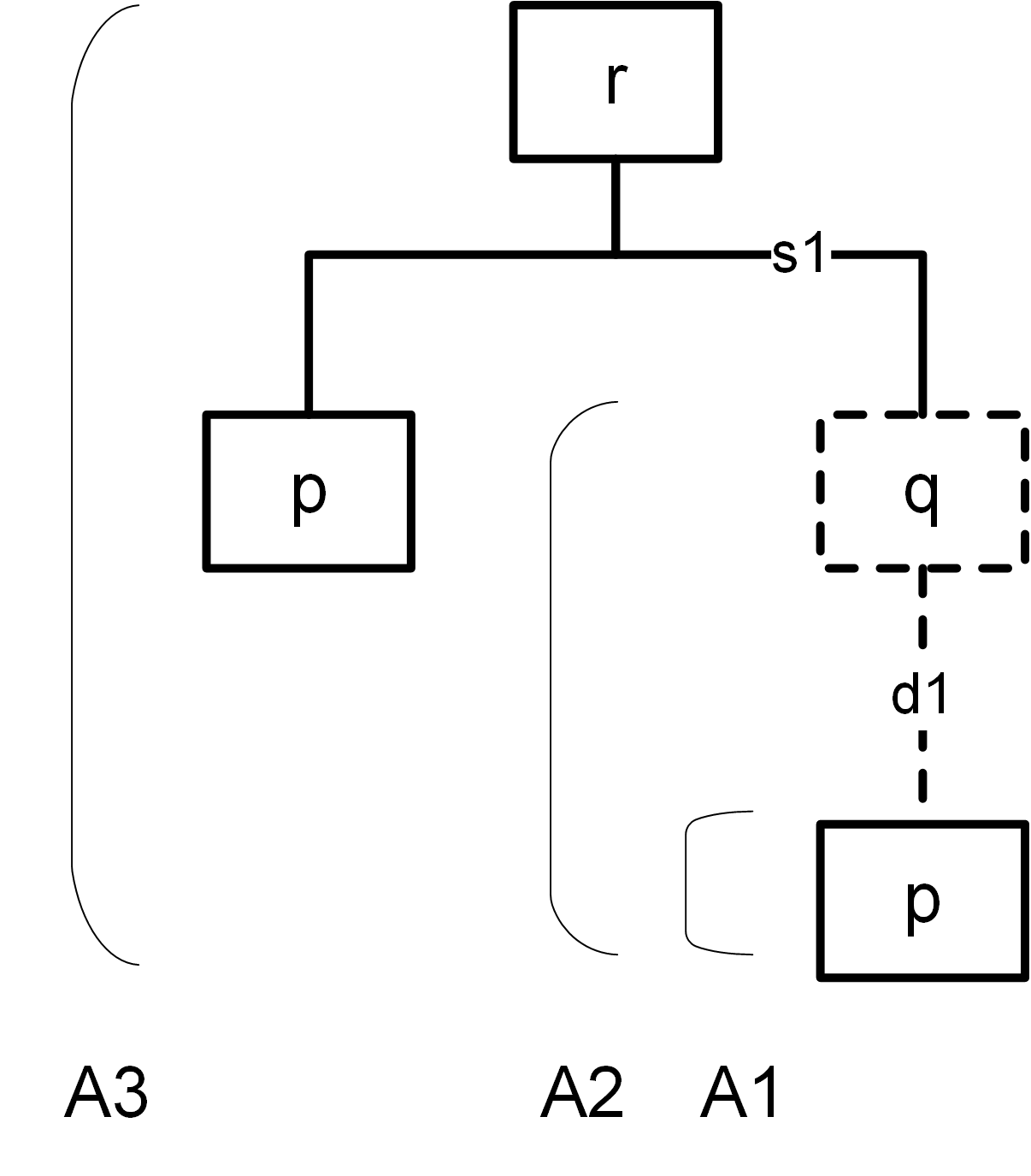}
	\caption{Argument $A_3$ from Example~\ref{exargument} with subarguments $A_1$ and $A_2$.}\label{ex:argument}
\end{figure}
%Formally the argument and its subarguments are written as follows:
%%%%
%\marginpar{CHECK Sanjay's NEW FIG}
%%%%
%\btab{l}
%$A_1$:~$p$ \\
%$A_2$:~$A_1 \Imp q$ \\
%$A_3$:~$A_1,A_2 \imp r$
%\etab
We have that
\btab{llllll}
$\mathtt{Prem}(A_3) =$ & $\{p\}$  &&& $\mathtt{DefRules}(A_3) =$ & $\{d_1\}$\\
$\mathtt{Conc}(A_3) =$ & $r$ &&& $\mathtt{Rules}(A_3) =$ & $\{s_1,d_1\}$\\
$\mathtt{Sub}(A_3) =$ & $\{A_1, A_2, A_3\}$ &&& $\mathtt{TopRule}(A_3) =$ & $s_1$ \\
$\mathtt{LDR}(A_3) =$ & $\{d_1\}$ &&&&
\etab
All of $A_1$, $A_2$ and $A_3$ are defeasible since $\mathtt{DefRules}(A_1) = \mathtt{DefRules}(A_2) = \mathtt{DefRules}(A_3) = \{d_1\}$.
\end{example}

\begin{figure}[ht]
\centering
		\includegraphics[scale=0.60]{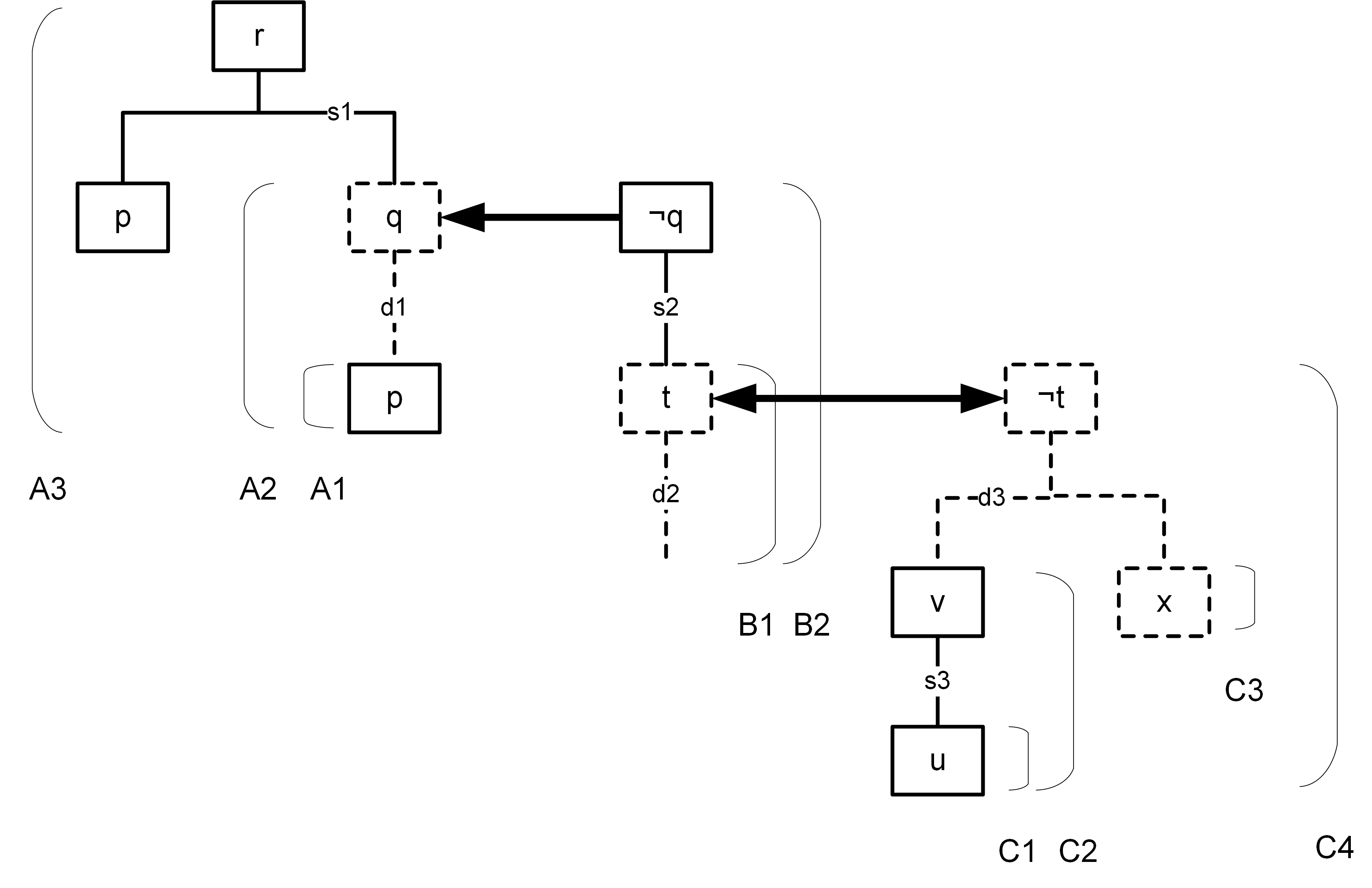}
	\caption{The arguments and attacks in the running example}\label{ex:arg2}
\end{figure}

In general,  \ASPIC\  has three ways of attack: on an argument's uncertain premises (undermining attack), on the conclusion of a defeasible rule (rebutting attack) and on a defeasible rule itself (undercutting attack). However, in this paper we only consider rebutting attack.
%since DeLP has no analogues to undercutting and undermining attack. Thus:
%
\begin{definition}\label{attacks}[\textbf{Rebutting Attack}] $A$ \emph{attacks} or \emph{rebuts}  $B$
iff $\mathtt{Conc}(A) = -\varphi$ for some  $B'$ $\in$
$\mathtt{Sub}(B)$ of the form $B''_1, \ldots, B''_n \Rightarrow
\varphi$.
\end{definition}
\begin{example}
In our running example argument  $A_3$ is rebutted on $A_2$ by an argument $B_2$ for $\neg q$:
\btab{l}
$B_1$:~$\Imp t$ \\
$B_2$:~$B_1\imp \neg q$
\etab
Note that $A_2$ does not in turn rebut $B_2$ on $B_2$, since $B_2$ has a strict top rule while the argument on which an argument is (directly) rebutted has to have a defeasible top rule. For the same reason $B_2$ can potentially only be rebutted on $B_1$. Our argumentation theory allows for such a rebuttal:
\btab{l}
$C_1$:~$u$ \\
$C_2$:~$C_1 \imp v$ \\
$C_3$:~$x$ \\
$C_4$:~$C_2,C_3 \Imp \neg t$
\etab
Note that $B_1$ in turn rebuts $C_4$, since $C_4$ has a defeasible top rule. All arguments and (direct) attacks in the example are displayed in Figure~\ref{ex:arg2}.
\end{example}
\citeN{c+a07} also consider a variant called `unrestricted rebut', which allows direct rebuttals on arguments with a strict top rule provided the attacked argument is defeasible:
 \begin{definition}\label{unr}[\textbf{Unrestricted Rebutting Attack}] $A$  \emph{u-rebuts}  $B$
iff $\mathtt{Conc}(A) = -\mathtt{Conc}(B')$ for some defeasible $B'$ $\in$ $\mathtt{Sub}(B)$.
\end{definition}
\begin{example}
In our running example this yields one additional rebutting relation, since $A_2$ u-rebuts $B_2$. Furthermore, argument $A_3$ can be potentially u-rebutted on its final conclusion $r$. However, $C_2$ cannot be u-rebutted, since it is not defeasible but strict.
\end{example}
Below we will assume Definition~\ref{attacks} of attack unless specified otherwise.

The \ASPIC\ counterpart of an abstract argumentation framework is a structured argumentation framework.
\begin{definition}\label{DefinitionStructuredAF}[\textbf{Structured Argumentation Frameworks}]
Let $AT$ be an \textit{argumentation theory} $(AS,\K)$.
A \textit{structured argumentation framework (\emph{SAF})} defined by $AT$ is a triple $\langle\A$, $\C$, $\preceq$ $\rangle$ where $\A$ is the set of all arguments on the basis of
%$\K$ over
$AS$, $\preceq$ is an ordering on $\A$, and $(X,Y) \in \C$ iff $X$ attacks $Y$.
\end{definition}
\begin{example}
In our running example $\A = \{A_1,A_2,A_3,B_1,B_2,C_1,C_2,C_3,C_4\}$, while $\C$ is such that $B_2$ attacks both $A_2$ and $A_3$, argument $C_4$ attacks both $B_1$ and $B_2$ and $B_1$ attacks $C_4$.
\end{example}
%
%%%%%%%%%%%%%%%%%%%%%%%%%%%%%%%%%%%%
%%%%%%%%%%%%%%%%%%%%%%%%%%%%%%%%%%%%
The attack relation tells us which arguments are in conflict with each other. If an argument $A$ \emph{successfully attacks}, i.e., \emph{defeats}, $B$, then $A$ can be used as a counter-argument to $B$. Whether a rebutting attack  succeeds as a defeat, depends on the argument ordering $\preceq$.  In the following definition  $A \prec B$ is defined as usual as $A \preceq B$ and $B \not\preceq A$.
\begin{definition}\label{defeat}[\textbf{Defeat}]. Argument $A$ \emph{defeats} argument $B$ if $A$ rebuts $B$ on $B'$ and $A \not\prec B'$.
\end{definition}
\begin{example}\label{ex:argord}
In our running example, the attack of $B_2$ on $A_2$ (and thereby on $A_3$) succeeds if $B_1 \not\prec A_2$. In that case $B_2$ strictly defeats both $A_2$ and $A_3$. If $B_1$ and $C_4$ are incomparable or of equal priority, then these two arguments defeat each other, while $C_4$ strictly defeats $B_2$. If $C_4 \prec B_1$ then $B_1$ strictly defeats $C_4$ while if $B_1 \prec C_4$ then $C_4$ strictly defeats both $B_1$ and $B_2$.
\end{example}
$AFs$ are then generated from $SAFs$ by letting the attacks from an $AF$ be the defeats from a $SAF$.
\begin{definition}[\textbf{\emph{AFs} corresponding to \emph{SAFs}}]
\label{def:af}
An \emph{abstract argumentation framework ($AF$) corresponding to a} $SAF$ = $\langle\A$, $\C$, $\preceq$   $\rangle$ (where $\C$ is \ASPIC's attack relation) is a pair $(\A, \mathit{attack})$ such that $\mathit{attack}$ is the defeat relation on $\A$ determined by $SAF$.
\end{definition}

A nonmonotonic consequence notion can then be defined as follows. Let $T \in \{$complete, preferred, grounded, stable$\}$ and let $\LA$ be from the $AT$ defining $SAF$. A wff $\varphi \in \LA$ is \emph{sceptically $T$-justified} in $SAF$ if $\varphi$ is the conclusion of a sceptically $T$-justified argument, and \emph{credulously $T$-justified} in $SAF$ if $\varphi$ is not sceptically $T$-justified and is the conclusion of a credulously $T$-justified argument.

\begin{example}\label{exargument:incon}
In our running example, if $B_2$ does not defeat $A_2$, then all extensions in any semantics contain $A_3$ so $r$ is sceptically justified. Let us next assume that $B_2$ defeats $A_2$.  Then if $C_4$ strictly defeats $B_1$, we have a unique extension in all semantics, namely, $\{A_1,A_2,A_3, C_1,C_2,C_3,C_4\}$. In both cases this yields that wff $r$ is sceptically justified. Alternatively, if $B_1$ strictly defeats $C_4$, then there again is a unique extension in all semantics, which now is $\{A_1,B_1,B_2,C_1,C_2,C_3\}$. Then $r$ is neither sceptically nor credulously justified. Finally, if $B_1$ and $C_4$ defeat each other, then the grounded extension is  $E =  \{A_1,C_1,C_2,C_3,C_4\}$ while there are two preferred extensions $E_1 = \{A_1,A_2,A_3,C_1,C_2,C_3,C_4\}$ and $E_2 = \{A_1,B_2,C_1,C_2,C_3\}$. So then $r$ is credulously but not sceptically justified in preferred semantics but is neither sceptically nor credulously justified in grounded semantics.
\end{example}

\subsection{Defeasible Logic Programming: \DLP}\label{sec:DeLP}

\hyphenation{pro-gra-mming}

Defeasible Logic Programming is a formalization of defeasible reasoning in which results of Logic Programming and Argumentation are combined.
\DLP\ has the declarative capability of representing knowledge in a language that extends the language of logic programming with the possibility of representing weak information in the form of \emph{defeasible rules}, and an argumentation-based inference mechanism for warranting conclusions.

%Unlike \ASPIC,
While \ASPIC\ in general abstracts from the logical language, \DLP chooses a logic-progra-mming language with  ``strong'' negation to represent knowledge in which the antecedents and consequent of a rule (strong or weak) are ground literals.
It is possible to employ in \DLP default negation, which is also known as negation as failure, but since this does not crucially change the analysis below, we will for simplicity ignore this extension here.
% grs - In \DLP\ default negation (also called negation as failure) can be used,} but since this does not crucially change the analysis below, we will ignore this extension here for simplicity.
As usual, rules written with free variables are schemes for all their ground instances.
Although \DLP and the instance of \ASPIC presented above are similar, they are not fully equivalent.
Elements in which \DLP and \ASPIC\ coincide are the predicate-logic literal language with strong negation, a set of indisputable facts, two sets of strict and defeasible rules, and a binary argument preference relation.
However, \DLP's definitions of argument, attack, and defeat are not equivalent to those of \ASPIC.
Moreover, a significant difference with \ASPIC\ is that \DLP, as defined by~\citeN{g+s04}, does not evaluate arguments by generating abstract argumentation frameworks.
Instead, \DLP's notion of \emph{warrant} is defined in terms of dialectical trees in a way that is similar to the argument game of grounded semantics but with some significant differences, as we will see below.

This section will introduce a description of  \DLP's features for knowledge representation (mainly taken from~\citeN{g+s14}); then, the details concerning its inference mechanism will be explained.
Although the work leading to the formalization of \DLP\ began in the early 90's~\cite{SChGarcia94a,SChGarcia94b} as an evolution of the work of~\citeN{s+l92}, its formalization was completed by~\citeN{GarciaPhD00} and finally published in~\citeN{g+s04,g+s14}.
Further developments can be found in the following related material: \cite{server07,TucatGS09,MartinezGS2012,explanation13,CohenGS16,g+s18}.

The knowledge representation language of \DLP\ is determined by a set of \emph{atoms}.
Atoms can be preceded by the  \emph{strong negation} symbol ``\no''.
Atoms that are not preceded by strong negation will also be called \emph{positive literals} and atoms preceded by strong negation will be called \emph{negative literals}, the term \emph{literal}, or sometimes \emph{objective literal}, will refer to either one.
A pair of literals involving a positive and a negative literal over the same atom are called \emph{complementary} or \emph{contradictory}.
For instance, ``$\no guilty$'' and ``$guilty$'' are two complementary literals.
A \emph{defeasible logic program},  abbreviated \dlp, is a set of facts, strict rules, and defeasible rules defined as follows:
\begin{itemize}
	\item[-]\emph{Facts} are ground (objective) literals, \eg $guilty$, $price(100)$, $\no close$.
	In \DLP\ facts are used for representing information that is considered to hold in the application domain. Hence,  as it will be explained below, a \dlp\ cannot contain two complementary facts.
	\item[-]\emph{Strict Rules} represent a relation between a ground literal $L_0$, or \emph{head} of the rule, and a set of ground literals $\{L_i\}_{i>0}$, or \emph{body} of the rule, and are denoted $\srule{L_0}{L_1, \ldots, L_n}$; strict rules correspond syntactically to \emph{basic rules} in Logic Programming~\cite{Lif96}. The use of the adjective `strict' emphasizes that the relation between the head and the body of the rule is such that if the body is accepted then the head must also be accepted.
	The examples of strict rules shown below can be understood as expressing that: \emph{someone who is guilty cannot be innocent},   \emph{cats are mammals}, and  \emph{if there are not many surfers  then there are few surfers}:
	\begin{center}
		\srule{\no innocent}{guilty}\\
		\srule{mammal}{cat}\\
		\srule{few\_surfers}{\no many\_surfers}\\
	\end{center}
	\item [-]\emph{Defeasible Rules} are used to represent a weaker connection between pieces of information, they are denoted $\drule{L_0}{L_1, \ldots, L_n}$, and like strict rules, the head of the rule $L_0$  is a ground literal and its body $\{L_i\}_{i\geq0}$ is a set of ground literals.
A defeasible rule with empty body is called a \emph{presumption}; sometimes we will also call the head of such a rule a presumption. Note that initially \citeN{g+s04} required defeasible rules to have non-empty bodies. Here and below we follow their extension in Section 6.2 of \DLP with presumptions. Unlike
	%%%
	%\marginpar{\henry{Doesn't this contradict your statements about strict closure?}}
	%%%
	strict rules, acceptance of the body of a defeasible rule does not always lead to the acceptance of the head.
	%In defeasible rules, literals on the body can be preceded with the default negation symbol ``\naf'' (see discussion below).
	Examples of defeasible rules follow. The first one represents that  \emph{usually, mosquitoes are not dangerous}, and the second says that \emph{reasons to believe mosquitoes are carrying dengue, justify the belief they are dangerous}:
	\begin{center}
		\drule{\no dangerous}{mosquito}\\
		\drule{dangerous}{mosquito,dengue}  \\
	\end{center}
\end{itemize}

Note that, from a syntactic point of view, strict and defeasible rules differ only in the symbol between the head and the body of the rule.
It is interesting to remark here that the representational choice between these two forms of relating the head and the body of a rule is ultimately a matter of context, sometimes a rule could change accordingly to the environment in which it is used; for instance, a rule that locally can be considered strict could become defeasible in a larger environment.
Defeasible rules allow to represent a weak connection between the body (antecedent) and the head (consequence) of the rule.
A defeasible rule \drule{H}{B} expresses that reasons to believe in $B$ provide a (defeasible) reason to believe in $H$.
As an example, consider an scenario where an agent has to decide how to spend the day.
Then, the defeasible rule ``\drule{nice}{waves}'' can represent that ``\emph{reasons to believe that there are big waves at the beach, is a reason to believe that it should be a nice day for surfing}''.
The connection between ``\emph{waves}'' and ``\emph{nice}'' is weak in the sense that there might be other reasons
such as ``\emph{normally, if it is raining it is not nice for surfing}'', represented as ``\drule{\no nice}{rain}'', that will lead to the contrary conclusion.
Suppose that today there are big waves and it is raining, then the acceptance of the body of the rule ``\drule{nice}{waves}'' does not lead directly to the acceptance of the head.

Nevertheless, strict rules establish a strict connection between body and head; thus, the rule  ``\srule{\no working}{vacation}'' represents the fact that in vacation an agent it is not working.
Then, as we will show below, due to this strict connection in \DLP\ if ``\emph{vacation}'' is accepted then ``\no~\emph{working}'' is also accepted.

Note that the symbols ``\drule{}{}'' and ``\srule{}{}'' denote meta-relations between a literal and a set of literals, and have no interaction with language symbols.
As in Logic Programming, strict and defeasible rules are not conditionals nor implications, they are inference rules.
%%%
%\marginpar{\henry{Rewritten on the basis of our discussions; please check.}}
%%%
%; consequently, it is not possible to apply contraposition to program rules, strict or defeasible. \grs{Therefore effecting a closure by ``strict rules'', as stated in the strict closure postulate~\cite{c+a07} has no meaning although some or all contrapositive rules of the rules in the program could be added explicitly by the knowledge engineer; if that is the case, the system will analyze the situation accordingly.
Consequently, strict rules do not automatically contrapose or (in \citeANP{c+a07}'s~\citeyear{c+a07} terms) `transpose'. In DeLP a knowledge engineer has to separately determine for each strict rule whether adding its transposition is appropriate for that rule. This is relevant since many positive results in the literature on satisfaction of \citeANP{c+a07}'s rationality postulates depend on the assumption that the set of strict rules is closed under transposition.

\begin{definition}[\textbf{Defeasible Logic Program}]
	A defeasible logic program (\dlp) is set of facts, rules and presumptions. However, when required, a \dlp\ is denoted \SD, to distinguish the subset \SSet\ of facts and strict rules and the subset \DD\ of defeasible rules and presumptions.
Moreover when we want to refer to just the facts in $\SSet$ we write $\SSet_f$ and for the strict rules we write $\SSet_s$. Naturally, $\SSet_f \cup \SSet_s = \SSet$.
\end{definition}

\begin{example} \label{ex.surf}
	\begin{center}
		\begin{small}
%			\begin{tabular}{l}
				$\SSet_{\ref{ex.surf}}$=$\left\{
				\begin{array}{lll}
				\facto{monday}\\
				\facto{cloudy}\\
				\facto{dry\_season}\\
				\facto{waves}\\
				\facto{grass\_grown}\\
				\facto{hire\_gardener}\\
				\facto{vacation}\\
				\srule{\no working}{vacation}\\
				\srule{few\_surfers}{\no many\_surfers}\\
				\srule{\no surf}{ill}\\
				\end{array}
				\right\}$
%			\end{tabular}
%			\begin{tabular}{l}
				   $\DD_{\ref{ex.surf}}$=$\left\{
				\begin{array}{lll}
				\drule{surf}{nice,spare\_time}\\
				\drule{nice}{waves}\\
				\drule{\no nice}{rain}\\
				\drule{rain}{cloudy}\\
				\drule{\no rain}{dry\_season}\\
				\drule{spare\_time}{\no busy}\\
				\drule{\no busy}{\no working}\\
				\drule{cold}{winter}\\
				\drule{working}{monday}\\
				\drule{busy}{yard\_work}\\
				\drule{yard\_work}{grass\_grown}\\
				\drule{\no yard\_work}{hire\_gardener}\\
				\drule{many\_surfers}{waves}\\
				\drule{\no many\_surfers}{monday}\\
				\end{array}
				\right\}$
%			\end{tabular}
		\end{small}
	\end{center}
\end{example}

\begin{definition}[\textbf{Defeasible Derivation}]\label{def.delp.derivation}
	Given a \dlp\ \SD,  a \emph{defeasible derivation} of a ground literal $L$ from \SD, denoted as $\SD \nm L$,  is a finite sequence $L_1,\ldots,L_n = L$ of ground literals such that for all $L_i (1 \leq i \leq n)$:
    $L_i \in \SSet$ or $L_i$ is a presumption in $\DD$; or there exists a rule $R_i$ in \SD\ (strict or defeasible) with head $L_i$ and body  $B_1,B_2,\ldots,B_m$ such that every literal  $B_j, 1 \leq j \leq m$,  of the body is an element $L_k$ already appearing in the sequence preceding $L_i$ ($k < i$).
If $L$ has a derivation that only uses facts and strict rules from \SSet\ and no defeasible rules, in this case we say that $L$  has  a \emph{strict derivation}.
\end{definition}

In the program \pair{\SSet_{\ref{ex.surf}}}{\DD_{\ref{ex.surf}}} shown in Example~\ref{ex.surf},
the literal \facto{surf} has a defeasible derivation:  \facto{vacation}, \facto{\no working},  \facto{\no busy}, \facto{spare\_time}, \facto{waves}, \facto{nice}, \facto{surf}, which contains two facts (\facto{vacation} and \facto{waves}), and the use of a strict rule
(\srule{\no working}{vacation}) and four defeasible rules.
Note that every fact of a \DLP\ has a  defeasible derivation; however, not every head of a rule has a derivation,
for instance, neither \facto{cold} nor \facto{\no surf} have a defeasible derivation.
Note that, \facto{\no working} has a strict derivation from $\SSet_{\ref{ex.surf}}$.
Note that literals that have a strict derivation must be facts or the head of a strict rules; however,
a literal can be the head of a strict rule and might have a defeasible derivation, but not a strict derivation.
For instance, \facto{few\_surfers}  has no strict derivation from $\SSet_{\ref{ex.surf}}$,
although  it has a defeasible derivation from \pair{\SSet_{\ref{ex.surf}}}{\DD_{\ref{ex.surf}}} that uses a defeasible rule for the derivation of \facto{\no many\_surfers}.

It is important to note that in \DLP\ the set \SSet\ is used to represent non-defeasible information, consequently it is required that the set be representationally coherent.
Therefore, for any program \SD\ we assume that \SSet\ is non-contradictory: no pair of contradictory literals can be derived from \SSet, \ie no strict derivation for complementary literals can be obtained from a \delp.
Saying that $\Pi$
is non-contradictory is equivalent to saying in \ASPIC\ that $\K$ is indirectly consistent relative to $\R_s$.

\begin{definition}\label{delp-arg}[\textbf{Argument}]
	Let \SD\ be a \dlp\ and $L$ a ground literal.
We say that \Arg\ is an \textbf{argument} for the conclusion $L$ from \SD, denoted \AL,
if \Arg\ is a set of defeasible rules (\Arg $\subseteq$\DD), such that:
	\ben
	\item there exists a defeasible derivation for $L$ from \SyA, and
	\item no pair of contradictory literals can be defeasibly derived from \SyA.
	\item $\A$ is minimal in that there is no proper subset of $\A$ satisfying conditions (1) and (2).
	\een
\end{definition}

Observe that although facts and strict rules are used in the defeasible derivation, the argument structure only mentions the defeasible rules, \ie facts and strict rules are not part of an argument.
Note also that unlike in Definition~\ref{arg} of \ASPIC-arguments, the set of defeasible rules of a \DLP argument has to be minimal and its set of `conclusions' has to be indirectly consistent. (Strictly speaking, the set of conclusions of a \DLP argument is not formally defined, but the set of all literals in the defeasible derivation corresponding to an argument can be seen as such.)
% -- grs \ale{Having additional constraints on \ASPIC\ arguments (as conditions 2 and 3 of Definition~\ref{delp-arg}) is formally explored by \citeN{w+p14}.} \rood{HENRY's PROPOSAL: delete this sentence and reference.}

Note that it could happen that a literal $L$ has a defeasible derivation from a \dlp\ but there is no argument for $L$ from that \dlp.
For instance, consider the \dlp\  \pair{\SSet_{\ref{ex.surf}}}{\DD_{\ref{ex.surf}}} of Example~\ref{ex.surf}, from the fact \facto{monday} and the defeasible rule \drule{working}{monday}, there is  a defeasible derivation for the literal \facto{working}.
However, note that there is a strict derivation for \facto{\no working}, and hence from the set $\SSet_{\ref{ex.surf}} \cup \{\drule{working}{monday}\}$ both literals  \facto{working} and  \facto{\no working} can be defeasibly derived, for that reason there is no argument for the literal \facto{working} from \pair{\SSet_{\ref{ex.surf}}}{\DD_{\ref{ex.surf}}}.
Consider $S = \{ \drule{nice}{waves},  \drule{\no busy}{\no working}\}\subseteq \DD_{\ref{ex.surf}}$.
Observe that  $S\cup \SSet_{\ref{ex.surf}}$ is non contradictory and allows for the defeasible derivation of \facto{nice}; however, $S$ is not an argument for \facto{nice} because it is not minimal.
Observe that  $\Arg_2 \subset S$ is an argument for \facto{nice}:
$\Arg_2= \{ \drule{nice}{waves}\}$.

Attack on arguments is in \DLP defined in terms of disagreement between literals.
Two literals $L$ and $Q$ are said to disagree in the context of the program \SD\ if the set $\SSet \cup \{L, Q\}$ is contradictory, \ie from  $\SSet \cup \{L, Q\}$ is possible to strictly derive a literal and its complementary.
For example, given $\SSet = \{(\srule{h}{a}), (\srule{\no h}{b}) \}$, the literals \facto{a} and \facto{b} disagree.
This notion of disagreement allows us to find direct and indirect conflicts between arguments.
 This is equivalent to saying in \ASPIC\ that $\K \cup  \{L,Q\}$ is indirectly inconsistent relative to $\R_s$.
Note that two complementary literals always disagree (\eg \facto{nice} and \facto{\no nice}).
Since for any program \SD\ it is required that \SSet\ be non-contradictory,
the disagreement cannot come from \SSet.

Given a \dlp\ \SD\ and two arguments \AL\ and \BQ\ obtained from it, if  \Barg\ $\subseteq$ \Arg\
then we say that \BQ\ is a \emph{subargument} of \AL\ and that \AL\ is the \emph{superargument} of \BQ\  (note that trivially every argument is a subargument/superargument of itself).

\begin{definition}[\textbf{Counterargument / Attack}]
In \DLP, an argument \BQ\ is a \emph{counterargument} for \AL\ at literal $P$,
if there exists a subargument \CP\ of \AL\ such that $P$ and $Q$ \emph{disagree}.
The literal $P$ is referred to as the \emph{counterargument point} and \CP\ as the \emph{disagreement subargument}. If \BQ\ is a counterargument for \AL, then we also say that \BQ\ \emph{attacks} \AL,
and that \BQ\ and \AL\ are in \emph{conflict}.
\end{definition}

Except for the disagreement check instead of a simple syntactic check for complementariness, this definition is similar to  Definition~\ref{attacks} of rebutting attack in \ASPIC\ in that an argument can attack a subargument of its target and does so on a specific point. On the other hand, unlike in \ASPIC, in \DLP the attacking point can be the consequent of a strict rule.
%, although such an attack will never exist without an attack on a previous defeasible step in the argument as well (since $\K \cup \R_s$ is assumed to be non-contradictory).
%%%%%%%%%%%%%%%%%%%%%%%%%%%%%%%%%%%%%%%%%%%%%%%%%%%%%%%%%%%%%%%%%%%%%%%%%%%%%%%%%%%%%%%%%%%%%%%%%%%%%%%%%
For instance, we include below Example 4 ``Married John'' from \citeN{c+a07} in terms of \DLP\ syntax.
%% \henry{Delete this example again?} We felt that the example was clarifying.
%
\begin{example}  \label{CaminadaAmgoud4}
Let $\SSet_{\ref{CaminadaAmgoud4}}$=$ \{ \facto{wr}, \facto{go}, 	\srule{\no hw}{b}, \srule{hw}{m},\}$  and $\DD_{\ref{CaminadaAmgoud4}}$=$ \{ \drule{m}{wr}, \drule{b}{go} \}$
%%Let S = {−→ wr;−→ go;b −→ ¬hw;m −→ hw} and D = {wr ?⇒ m;go ?⇒ b}
with:
\facto{wr} = ``John wears something that looks like a wedding ring'',
\facto{go} = ``John often goes out until late with his friends'',
\facto{hw} = ``John has a wife'',  \facto{b} = ``John is a bachelor'',
\facto{m} = ``John is married''.
The following arguments can be constructed:
\btab{llllll}
$\Arg_1$: & $\langle \{\ \}, wr\} \rangle$ &&& $\Arg_2$: & $\langle \{\ \}, go \rangle$ \\
$\Arg_3$: & $\langle \{\drule{m}{wr}\}, m\} \rangle$ &&& $\Arg_4$: & $\langle \{\drule{b}{go}\}, b \rangle$ \\
$\Arg_5$: & $\langle \{\drule{m}{wr}\}, hw\} \rangle$ &&& $\Arg_6$: & $\langle \{\drule{b}{go}\}, \no hw \rangle$ \\
\etab
In \DLP\ arguments $\Arg_1$ and $\Arg_2$ have no defeaters; argument $\Arg_5$ defeats $\Arg_6$ and vice versa;
and  argument $\Arg_3$ defeats $\Arg_4$ and vice versa. Note  also that argument $\Arg_3$ defeats $\Arg_6$
and argument $\Arg_4$ defeats $\Arg_5$.  Consequentely,  \facto{b} = ``John is a bachelor'' and
\facto{m} = ``John is married''  are not warranted (justified).
\end{example}

%%%%%%%%%%%%%%%%%%%%%%%%%%%%%%%%%%%%%%%%%%%%%%%%%%%%%%%%%%%%%%%%%%%%%%%%%%%%%%%%%%%%%%%%%%%%%%%%%%%%%%

As a further example, consider the \dlp\ \pair{\SSet_{\ref{ex.surf}}}{\DD_{\ref{ex.surf}}} of Example~\ref{ex.surf} and the sets
\Arga = \{(\drule{\no nice}{rain}) ; (\drule{rain}{cloudy})\},
\Argb = \{\drule{nice}{waves}\}, \Argc = \{\drule{rain}{cloudy}\}, \Argd = \{\drule{\no rain}{dry\_season}\}.
Then, \Ar{\Arg_1}{\no nice} is a counterargument for \Ar{\Arg_2}{nice} and vice versa because
in this particular case the conclusion of both arguments disagree.
As another example,  \Ar{\Arg_4}{\no rain} is a counterargument for \Ar{\Arg_1}{\no nice} at the counterargument point $rain$ and \Ar{\Arg_3}{rain} is the disagreement subargument.
Note that in \DLP\ a counter-argument for an argument \Arg\ is also a counter-argument for any super-argument of \Arg.
Also note that in \DLP\ there is no possible counterargument for a claim having a strict derivation, see~\citeN{g+s04} for the proof.
Observe that  from \pair{\SSet_{\ref{ex.surf}}}{\DD_{\ref{ex.surf}}} there is a strict derivation for \facto{\no working}, however, although there is a derivation for \facto{working}, no argument for \facto{working} can exists, and hence, no counter-argument for \facto{\no working}.

The argument comparison criterion is modular in \DLP; hence, it is possible to use any preference criterion established over the set of arguments (see~\citeN{g+s14} for details and~\citeN{TezeGGS15} for an application).
This allows the user to select the most appropriate criterion for the application domain that is being represented.
For the rest of the presentation we will assume an abstract preference criterion $\prec$ of strict comparison on the set of arguments, where $A \prec B$ means that argument $B$ is strictly better than argument $A$.

\begin{definition}[\textbf{Defeaters}]
Argument $\langle \A_1, L_1 \rangle$ is a \emph{proper defeater} of argument $\langle \A_2, L_2 \rangle$ iff there exists a subargument $\langle \A, L \rangle$ of $\langle \A_2, L_2 \rangle$ such that $\langle \A_1, L_1 \rangle$ attacks  $\langle \A_2, L_2 \rangle$ at literal $L$ and $\langle \A, L \rangle \prec \langle \A_1, L_1 \rangle$.
Argument $\langle \A_1, L_1 \rangle$ is a \emph{blocking defeater} of argument $\langle \A_2, L_2 \rangle$ iff there exists a subargument $\langle \A, L \rangle$ of $\langle \A_2, L_2 \rangle$ such that $\langle \A_1, L_1 \rangle$ attacks  $\langle \A_2, L_2 \rangle$ at literal $L$ and $\langle \A, L \rangle \not\prec \langle \A_1, L_1 \rangle$ and $\langle \A_1, L_1 \rangle \not\prec \langle \A, L \rangle$.
Argument $A$ is a \emph{defeater} of argument $B$ iff $A$ is a proper or a blocking defeater of $B$.
\end{definition}
In the context of program \pair{\SSet_{\ref{ex.surf}}}{\DD_{\ref{ex.surf}}},
\Ar{\Arg_4}{\no rain} is a counterargument for \Ar{\Arg_1}{\no nice} at the counterargument point $rain$ and \Ar{\Arg_3}{rain} is the disagreement subargument; therefore, \Argd\ is compared with \Argc\ to determine if it is a defeater.

To facilitate comparison of both approaches, in the rest of the paper a \DLP\ strict rule \srule{a}{b} can also be denoted using \ASPIC\ notation as $b \rightarrow a$, and defeasible rule \drule{a}{b} as $b \Imp a$.
Also, a \dlp\ \SD\ can be denoted $(\K,\R_s,\R_d)$ assuming \SSet = $\K \cup \R_s$ and $\DD = \R_d$.

The notions of proper and blocking defeater are not equivalent to the \ASPIC\ notions of strict and weak defeater (see Section~\ref{sec:afs}). Consider the following example:

\begin{example}\label{excrossover}
Consider a \dlp\ with $\K = \{\facto{q},\facto{s}\}$, $\R_s = \emptyset$ and $\R_d = \{\drule{p}{};~\drule{r}{p,q};~\drule{\neg r}{};~\drule{\neg p}{\neg r, s}\}$.
Then we have the following \DLP arguments:
	\btab{llllll}
	$\Arg_1$: & $\langle \{\drule{p}{}\}, p\} \rangle$ &&& $\Arg_2$: & $\langle \{\drule{p}{};~\drule{r}{p,q}\}, r \rangle$ \\
	$\BArg_1$: & $\langle \{\drule{\neg r}{}\}, \neg r \rangle$ &&& $\BArg_2$: & $\langle \{\drule{\neg r}{};~\drule{\neg p}{\neg r,s} \}, \neg p \rangle$ \\
	\etab
	And let $\Arg_1 \prec \BArg_2$ and $\BArg_1 \prec \Arg_2$ (a preference based on strict specificity). Note that $\Arg_1$ is a subargument of $\Arg_2$ and $\BArg_1$ is a subargument of $\BArg_2$. Then, $\Arg_2$ and $\BArg_2$ are proper defeaters of each other, while the \ASPIC\ relation of strict defeat is asymmetric. Note also that $\Arg_2$ and $\BArg_2$ weakly defeat each other while they are not blocking defeaters of each other.
\end{example}

An \emph{argumentation line}  for an argument \ALa\ is a sequence of arguments from a \dlp, denoted
$\Aline\,=[\ALa,\,\ALb,\,\ALc,\,\ldots]$,
where each element of the sequence \ALi, $i>1$, is a defeater of its predecessor \ALiAnt.
The first element, \ALa, becomes a \emph{supporting} argument for the conclusion \La, \ALb\ an \emph{interfering} argument, \ALc\ a supporting argument, \ALd\ an interfering one,  continuing in that manner.
Thus, an argumentation line can be split into two disjoint sets:
$\Sline\,=\,\{\ALa, \ALc, \ALe, \ldots\}$ of supporting
arguments for the conclusion \La, and \Iline\,=\,\{\ALb,\,\ALd, $\ldots$\} of interfering arguments for \La.

\begin{definition}
Given a program \SD, a set of arguments $\{\ALi\}_{i=1}^{k}$ is concordant if it is not possible to have a defeasible derivation for a pair of contradictory literals from the set $\SSet \cup \bigcup_{i=1}^{k} \Argi$.
\end{definition}

\begin{definition}\label{accline}[\textbf{Acceptable Argumentation Line}]\footnote{\citeANP{g+s04}~\citeyear{g+s04}'s definition  of acceptable argumentation line has been modified in a 2014 work~\cite{g+s14} and recently in an unpublished paper correcting some unsuitable behavior in particular cases pointed out by Henry Prakken (see Example~\ref{ex2wrong}).}
An argumentation line \Aline\,=\,$[\ALa,  \ldots \ALn ]$ from a \dlp\ \SD\ is \emph{acceptable}  if and only if:
\begin{enumerate}\itemsep0pt
	\item \Aline\ is a finite sequence.
	\item The set \Sline\ of supporting arguments (resp. \Iline)  is concordant.
	\item  No argument \ALk\ in \Aline\ is a subargument of an argument \ALi\ appearing earlier in \Aline, $i<k$.
	\item  For all $i$, such that  \ALi\ is a blocking defeater for \ALiAnt,
	if \ALiProx\ exists, then \ALiProx\ is a proper defeater for \ALi.
\end{enumerate}
\end{definition}

Given a program,  there can be more than one argumentation line starting with the same argument \AL.
Therefore, analyzing a single acceptable argumentation line for \AL\ will not be  enough to determine whether \AL\ is an undefeated argument.
In the general situation, there might be several defeaters \BQa, \BQb, $\ldots$, \BQk\  for  \ALa, and for each defeater \BQi\  there could be in turn several defeaters; thus, a  tree structure is defined which is called a dialectical tree.
In this tree, the root is labeled with \AL\ and every  node (except the root) represents a defeater (proper or blocking) of its parent.
Each branch in the tree, \ie each path from a leaf to the root, corresponds to a different acceptable argumentation line.

\begin{definition}[\textbf{Dialectical Trees}]
Let  \ALa\ be an argument obtained from a \DLP-program \PP,
a \emph{dialectical tree} for \ALa\  from \PP\ is  denoted \Tree{\ALa} and  is constructed as follows:
\begin{enumerate}\itemsep 2pt
	\item The root of the tree is labeled with  \ALa.
	\item Let $N$ be a node labeled  \ALn, and
	$[\ALa, \ldots, \ALn]$ be the sequence of labels of the path
	from the root to $N$. Let \{\BQa, \BQb, $\ldots$ ,\BQk\} be the set of all
	the defeaters for \ALn\ from \PP.
	For each defeater \BQi\ $(1\leq i \leq k)$, such that
	the argumentation line $\Aline'= [\ALa, \ldots, \ALn, \BQi]$ is acceptable,
	the node $N$ has a child $N_i$ labeled  \BQi.
	If there is no defeater for \ALn\ or there is no \BQi\
	such that $\Aline'$ is acceptable, then $N$ is a leaf.
\end{enumerate}
\end{definition}

A dialectical tree provides a useful structure for considering all possible acceptable argumentation lines that can be generated for deciding whether the starting argument is defeated.
Given a literal $L$ and an argument \AL, to decide whether the literal $L$ is warranted, every node in the dialectical tree \Tree{\AL} is  recursively marked as \Dnode\ (\emph{defeated}) or \Unode\ (\emph{undefeated}), obtaining a marked dialectical tree \MTree{\AL}.
Nodes are marked by a bottom-up procedure that starts marking all leaves in \MTree{\AL}
as \Unode s. Then, for each inner node  \BQ\ of \MTree{\AL}, either: \\[4pt]
(a) \BQ\ will be marked as \Unode\  iff every child of \BQ\ is marked as  \Dnode, or \\[4pt]
(b) \BQ\ will be marked as \Dnode\   iff it has at least a child marked as \Unode.
\smallskip

This marking procedure provides an effective way of determining if a \DLP-query $L$ is warranted.
It is important to note that  given a \DLP-query $L$, there can be several arguments that support $L$;
therefore,  $L$ will be warranted if there exists at least one argument \Arg\ for $L$ such that the root of a  dialectical tree for \AL\ is marked as \Unode.
Given an argument \AL\ obtained from a program \PP, we will write  $\mathit{Mark}(\MTree{\AL})=U$ to denote
that the root of \MTree{\AL} is marked as \Unode;
otherwise we will write $\mathit{Mark}(\MTree{\AL})=D$ (if the root of \MTree{\AL} is marked as \Dnode).
Thus, we can define warrant in terms of the marking procedure $\mathit{Mark}$:

\begin{definition}[\textbf{Warrant}]
	Let \SD\ be a \dlp\ and $L$ a ground literal.
	We say that  $L$ is \emph{warranted} from \SD\ if there exist at least one argument \AL\ from \SD,
	such that $\mathit{Mark}(\MTree{\AL})=U$, we also say that \MTree{\AL} \emph{warrants} $L$ and that \Arg is a \emph{warrant} for $L$.
When no such argument exists the literal $L$ is said to be \emph{unwarranted}.
\end{definition}

Each acceptable argumentation line can be seen as a two-player argument game like the grounded game except that the rules of the game are given by the conditions of Definition~\ref{accline} on acceptable argument lines. Thus there is an equivalence between the above definition of warrant and the notion of a winning strategy for the proponent in the corresponding argument game. Given a dialectical tree in which the root is labelled $U$, the proponent in the game has a winning strategy by for each defeater moved by the opponent picking a reply from the tree that is labelled $U$. Conversely, if the proponent has a winning strategy in a game for argument \Arg, then \Arg will clearly have to be labelled $U$ in its dialectical tree, since this tree contains the winning strategy as a subtree that contains all legal defeaters of any supporting argument in the tree. This equivalence will be exploited below in Section~\ref{groundedDeLP} in the proposal to consider a version of \DLP where the conclusions will be obtained through grounded semantics, we distinguish this particular system by denoting it as \DLPgr.

%%%%%%%%%%%%%%%%%%%%%%%%%%%%%%%%%%%%%%%%%%%%%%%%%%%%%%%

\section{Comparing the Argument Definitions}\label{comparing}

In this section we compare the argument definitions of \DLP and \ASPIC.
% grs ---- Since in the previous section we argued that it would be interesting to formulate a version \DLP with grounded semantics by adopting the argument game for this semantics, we will in this section consider a thus reformulated version of \DLP.
% grs ---- This will also facilitate the comparison with \ASPIC.
At first sight, the deduction nature of \DLP arguments would seem to allow a straightforward many-to-one mapping onto \ASPIC\ arguments in that \DLP arguments would capture one possible ordering of the inferences in an \ASPIC\ arguments. A mapping of this kind  was by \citeN{hp10aspicJAC} established between the arguments of  assumption-based argumentation and \ASPIC\ arguments. However, two  features of \DLP arguments prevent a straightforward mapping onto  \ASPIC\ arguments: the minimality requirement and the consistency requirement. We discuss both requirements in turn.

\subsection{On Rationality Postulates}\label{Rationality.Postulates}

In this and the following sections we will report several positive and negative results on satisfaction of the rationality postulates of \citeN{c+a07}. We now make some introductory remarks on
%%%
%\marginpar{HP: subsection added as agreed.}
%%%
these postulates, in order to put the later results into perspective. While it is hard to disagree that the postulates of direct consistency and closure under subarguments should be satisfied, this is different for indirect consistency and closure under strict rules.\footnote{See similar discussions in formal epistemology on whether justified beliefs should be classically consistent and closed under deduction \cite{nel00}.} Various positions can be adopted. One position (which is the one of Caminada \& Amgoud)  is that strict closure and indirect consistency should always be satisfied, given the intuitive reading of strict rules $S$ strictly implies $p$ as ``If $S$ then always, or without exception, $p$''.

Another position (which is the one of the first and third author of this paper), is that these properties are only desirable for sets of arguments that are not attackable (for instance, in \DLP or \ASPIC\ sets of arguments that only use facts and strict rules). By contrast, if an antecedent of a strict rule is provided by a defeasible rule, then it may be reasonable to not accept the consequent of the strict rule even if all its antecedents are accepted. This is one reason why a knowledge engineer in \DLP has to determine separately for each strict rule whether it transposes (cf.\ Section~\ref{sec:DeLP} above).

A third position (adopted by the second author of this paper in \cite{hp16}) is that what is decisive is the properties of the argument ordering. \citeN{hp10aspicJAC} defined when an argument is ``reasonable'' (in a technical sense) and showed that (together with indirect consistency of the necessary part of the knowledge base and closure of the strict rules under contraposition or transposition) both strict closure and indirect consistency are satisfied for \ASPIC\ if the argument ordering is `reasonable'.  \citeN{m+p13aij} showed the same for a weaker definition of reasonable argument orderings. \citeN{hp16} agrees with the second position that strict closure should only hold in general for sets of arguments that are not attackable. He then argues that whether strict closure and indirect consistency should hold for other cases depends on whether it makes sense to require that the argument ordering is `reasonable', and, so he argues, this depends on the nature of the knowledge and inference rules.

With this in mind, in the remainder of this paper several results on (non-)satisfaction of strict closure and indirect consistency will be reported in a neutral way, without taking a stance on whether these results are good or bad for the investigated system(s). This posture is to leave the door open for more research on this crucial topic.

\subsection{Minimality}\label{sec:minimality}

\DLP requires arguments to be subset-minimal in their sets of defeasible rules. The general \ASPIC\ framework imposes no explicit minimality conditions on arguments, although the definition of an \ASPIC\ argument is such that it cannot contain `unused'  premises or rules: if an argument $A$ contains inferences, then all premises in $\mathtt{Prem}(A)$ and all rules in $\mathtt{Rules}(A)$ are used in at least one inference. In various publications the addition of minimality conditions on arguments has been studied. \citeN{m+p13aij} study a minimality requirement on the set of premises in order to establish relations with classical argumentation as studied by \citeN{g+h11}. \DLP does not impose minimality of premise sets (which in \DLP are the facts used in an argument).
%For example, with a defeasible rule $p \Imp q$ and a strict rule $p, r \imp q$ and $\K = \{p,r\}$, both an argument for $q$ using $p \Imp q$ and an argument for $q$ using $p, r \imp q$ are constructible.
It can even be the case that a \DLP argument which is minimal in its set of defeasible rules is non-minimal in its sets of premises or strict rules. Consider:
\btab{ll}
$A$: & $\langle \{f_1 \Imp p;~p,f_2 \imp q\}, q\rangle$ \\
$B$: & $\langle \{f_1 \Imp p;~f_1 \Imp r;~p,r \Imp q\}, q\rangle$
\etab
where $\K = \{f_1,f_2\}$. Argument $A$ is minimal in its set of defeasible rules but $B$ is minimal in its sets of premises and strict rules.
%For these reasons we will not further study minimality constraints on the sets of premises and strict rules of arguments.
The \ASPIC\ definition of an argument allows arguments that are non-minimal in their set of defeasible rules.
The just-given example illustrates this, since \ASPIC\ counterparts of both $A$ and $B$ can be constructed.

Requiring arguments to be minimal in their set of defeasible rules makes sense on the assumption that arguments with a non-minimal set of defeasible rules can never be stronger than a minimal version with the same conclusion (this assumption does not hold in general for \ASPIC).
For \DLP this assumption is reasonable, since defeasible rules are the only fallible elements in a \DLP argument.
Given this assumption, the above example shows that if arguments are required to be minimal in their set of defeasible rules, they cannot be required to be also minimal in their sets of premises and/or strict rules.

%Finally, the minimality constraint on the set of defeasible rules of an argument implies that \DLP arguments are non-circular in that no proper subargument of an argument can have the same conclusion. {\bf *** CHECK, since defeasible derivations don't need to be minimal. ***} By contrast, in \ASPIC\ as generally defined, arguments are not required to be non-circular, although versions have been studied with this requirement, viz.\ in \citeN{g+p16} and \citeN{hp18kr}. It should be noted that if strict-rule application is made explicit in a \DLP argument, then DeP arguments can be circular, since \DLP has no minimality requirement on the set of strict rules of an argument.
%

\subsection{Consistency}\label{conssec}

\DLP arguments have to be consistent in that no pair of complementary literals should be derivable from the set of all facts and strict rules of the program plus the defeasible rules used in the argument. In \ASPIC\ this would amount to saying that the set $\mathtt{Conc}(\mathtt{Sub}(A)) \cup \K$ is indirectly consistent (since as noted above, all rules of an \ASPIC\ argument are used to derive conclusions). Only one publication on \ASPIC\ has studied the same constraint, namely, \citeN{hp16}. However, in that paper the constraint was combined with a different notion of rebutting attack, in order to capture forms of probabilistic reasoning. \citeN{w+p14} study a slightly weaker constraint, namely, that $\mathtt{Sub}(A)$ is indirectly consistent.
%They prove that this does not affect the results on the rationality postulates. They also prove that for instantiations with classical-logic argumentation it does not affect the extensions under the assumption that an \ASPIC\ argument with a non-minimal set of premises can never be stronger than  a minimal version with the same conclusion. But they give counterexamples for the general case with defeasible rules. most notably in \cite{g+p16}.

The \DLP consistency constraint is intuitively appealling. However, \citeN{w+p14} remark that  their slightly weaker constraint for \ASPIC\ arguments induces counterexamples to indirect consistency. A very similar example can be constructed in \DLP.
\begin{example}\label{counterex2}
Consider a \dlp\ with $\SSet_f = \{f_1,f_2\}$,  $\SSet_s = \{\srule{r}{p,q};~\srule{\neg q}{p, \neg r}\}$ and $\DD = \{\drule{p}{f_1};~\drule{\neg r}{f_2};~\drule{q}{p}\}$. This enables the following \DLP arguments:
\btab{ll}
$A_1$: & $\langle \{\drule{p}{f_1}\},p \rangle$ \\
$A_2$: & $\langle \{\drule{p}{f_1};~\drule{q}{p}\},q \rangle$ \\
$A_3$: &  $\langle \{\drule{p}{f_1};~\drule{q}{p};\srule{r}{p,q}\},r \rangle$ \\
$A_4$: & $\langle \{\drule{\neg r}{f_2}\}, \neg r \rangle$ \\
$A_5$: & $\langle \{\drule{\neg r}{f_2};~\drule{p}{f_1};\srule{\neg q}{p, \neg r}\}, \neg q\rangle$ \\
%$A_6$: & $\langle \{f_1 \Imp p;~p \Imp q;~f_2 \Imp \neg r;~q,\neg r \imp \neg p\},\neg p \rangle$
\etab
%Note that $\mathtt{Conc}(\mathtt{Sub}(A_6))$ is indirectly inconsistent, so $A_6$ is not constructible in DeLP.
%If in \ASPIC\ it is also regard as unconstructible, then the attack relations in \ASPIC\ are that $A_3$ attacks $A_4$ while $A_5$ attacks $A_2$. However, in
In  \DLP both  $A_2$ and $A_5$ and $A_3$ and $A_4$ attack each other. If these conflicts are resolved with a last-link ordering on arguments as defined by \citeN{m+p13aij}, then the following sets of rules have to be compared:
\btab{l}
$\mathtt{LDR}(A_2) = \{p \Imp q\}$ with
$\mathtt{LDR}(A_5) = \{f_1 \Imp p;~f_2 \Imp \neg r\}$\\
$\mathtt{LDR}(A_3) = \{f_1 \Imp p;~p \Imp q\}$ with
$\mathtt{LDR}(A_4) = \{f_2 \Imp \neg r\}$
\etab
If the rules are in increasing order of priority ordered as $f_1 \Imp p < f_2 \Imp \neg r < p \Imp q$ then with the last-link ordering, by comparing sets on their minimal elements, we obtain that $A_3 \prec A_4$ and $A_5 \prec A_2$.
%Then in \ASPIC\ neither of the two attack relations results in defeat, so the grounded extension contains all five arguments and thus violates both strict closure and indirect consistency. By contrast, in
 Then, in \DLP, the attacks of $A_2$ on $A_5$ and $A_4$ on $A_3$ succeed as proper defeats, so $A_1,A_2$ and $A_4$ are warranted while $A_3$ and $A_5$ are not warranted. Thus, the set of warranted arguments is not strictly closed and not indirectly consistent.
\end{example}
We can  conclude from this example that  \DLP's strong consistency requirement on arguments does not in general suffice for satisfying strict closure and indirect consistency.
 %%% END \henry
% 

%%%%%%%%%%%%%%%%%%%%%%%%%%%%%%%%%%%%%%%%%%%%%%%%%%%%%%%
\section{Comparing the attack relations}\label{attacksec}

In order to compare the attack relations of DeLP and \ASPIC, we first define an \ASPIC\ counterpart of \DLP rebuttal.

\begin{definition}\label{dlp-attacks}[\textbf{dlp-rebutting attack}] $A$ \emph{dlp-rebuts}  $B$ iff  for some  $B'$ $\in$ $\mathtt{Sub}(B)$ it holds that $\mathtt{Conc}(A) \cup \mathtt{Conc}(B) \cup \K$ is indirectly inconsistent.
\end{definition}
It is easy to verify for \ASPIC\ that rebut implies unrestricted rebut and unrestricted rebut implies dlp-rebut. Counterexamples to the converse implications can easily be constructed.

Next we address the question whether adopting dlp-rebut (but not \DLP's strong consistency condition) could improve \ASPIC. So in the remainder of this section we assume all definitions of \ASPIC\ except that rebutting attack is replaced with dlp-rebutting attack. In particular, we do for now not require that arguments are non-contradictory in the sense of Definition~\ref{delp-arg}(2).

For complete, preferred and stable semantics the answer to our question arguably is negative, as can be shown with the following example, inspired by an example of \citeN{c+w11}, who read it as `any two of three persons can ride on a tandem together, but they cannot ride the tanden together all three of them'.

%\begin{example}\label{counterex} {\bf *** aanpassen aan \ASPIC\ notatie ***}\\
%Consider a dlp with $\K = \{f_1,f_2,f_3\}$ and  $\R_s = \{p,q \imp \neg r;~p,r \imp \neg q;~q,r \imp \neg p\}$ and $\R_d$ consisting of the defeasible rules in the following arguments:
%\btab{ll}
%$A$: & $\langle \{f_1 \Imp p\},p \rangle$ \\
%$B$: & $\langle \{f_2 \Imp q\},q \rangle$ \\
%$C$: & $\langle \{f_3 \Imp r\},r \rangle$ \\
%$A$+$B$: & $\langle \{f_1 \Imp p, f_2 \Imp q\},\neg r \rangle$ \\
%$A$+$C$: & $\langle \{f_1 \Imp p, f_3 \Imp r\},\neg q \rangle$ \\
%$B$+$C$: & $\langle \{f_2 \Imp q, f_3 \Imp r\},\neg p \rangle$
%\etab

\begin{example}\label{counterex}
Consider an \ASPIC\ AT  with $\K = \{f_1,f_2,f_3\}$ and  $\R_s = \{p,q \imp \neg r;~p,r \imp \neg q;~q,r \imp \neg p\}$ and $\R_d$ consisting of the defeasible rules in the following arguments:
\btab{ll}
$A$: & $f_1 \Imp p$ \\
$B$: & $f_2 \Imp q$ \\
$C$: & $f_3 \Imp r$ \\
$A+B$: & $A,B \imp \neg r$ \\
$A+C$: & $A,C \imp \neg q$ \\
$B+C$: & $B,C \imp \neg p$
\etab
With DeLP rebut, $A+B$ and $C$ rebut each other, $A+C$ and $B$ rebut each other and $B+C$ and $A$ rebut each other. If all arguments are incomparable in the argument ordering, then all these attacks succeed as defeats, so there exists an admissible set containing all of $A,B$ and $C$, which violates strict closure and indirect consistency. Note that $\R_s$ is closed under transposition. On the other hand, the grounded extension is $\{f_1,f_2,f_3\}$, which is strictly closed and indirectly consistent.
\end{example}
This example shows that if satisfying strict closure and indirect consistency is regarded as desirable, then adopting dlp-rebut in \ASPIC\ is not in general an improvement but may be an improvement in special cases. The example also shows that adopting dlp-rebut affects the set of extensions in at least complete, preferred and stable semantics even if the strict rules are closed under transposition.
%{\bf *** What about grounded semantics? And inclusion relations? Probably not, because of nonmonotonicity. ****}

\citeN{cmo14} prove for unrestricted rebut (see Definition~\ref{unr} above) that for a limited case with a total preference ordering on the set of defeasible rules and a weakest- or last-link argument ordering, the grounded extension satisfies both strict closure and indirect consistency (under the assumption that the set of strict rules is closed under transposition). Since unrestricted rebut and dlp-rebut are similar, it is interesting to see if similar results can be obtained for dlp-rebut. We first  investigate this for the so-called \emph{simple argument ordering}, which is such that $A \preceq B$ iff $A$ is defeasible and $B$ is strict.
%The proofs below follow those of \cite{cmo14} as closely as possible.

Direct consistency can be easily shown on the assumption that $\K$ is indirectly consistent.
\begin{proposition}~\label{direct-consistency}
Suppose attack in \ASPIC\ is dlp-rebut, $\K$ is indirectly consistent and the argument ordering is simple. Then for any $AF$ corresponding to a $SAF$ with grounded extension $E$, it holds that there is no $\varphi$ such that both $\varphi$ and $\neg \varphi$ are in $\mathtt{Concs}(E)$.
\end{proposition}
\begin{proof}
Suppose for contradiction $\varphi, \neg \varphi \in \mathtt{Conc}(E)$. Then there exist two arguments $A$ and $B$ in $E$ such that $\mathtt{Conc}(A) = \varphi$ and $\mathtt{Conc}(B) = \neg \varphi$. Since $\K$ is assumed to be indirectly consistent, at least one of $A$ and $B$ is defeasible. Assume without loss of generality that $B$ is defeasible. Then  $A \not\prec B$ so $A$ defeats $B$. But then $E$ is not conflict free.
\end{proof}

However, unlike in the case studied by \citeN{cmo14} there are counterexamples to strict closure and indirect consistency for \ASPIC\ with dlp-rebut even if $\K$ is indirectly consistent and $\R_s$ is closed under transposition.
\begin{example}
Let $\K = \{t\}$ and $\R_d = \{\Imp a_1,\Imp a_2, \Imp q\}$ while $\R_s$ consists of the following rules:

\btab{lllll}
$a_1,a_2 \imp p$ & $p, q \imp r$ & $p,q \imp \neg r$ & $t,r \imp s$ & $t,r \imp \neg s$\\
$a_1,\neg p \imp \neg a_2$ & $p, \neg r \imp \neg q$ & $p,r \imp \neg q$ &  $t, \neg s \imp \neg r$ & $t, s \imp \neg r$\\
$a_2,\neg p \imp \neg a_1$ & $q, \neg r \imp \neg p$ &		$q,r \imp \neg p$  & $r, \neg s \imp \neg t$ & $r, s \imp \neg t$\\
&&&& \\
$t,\neg r \imp u$ & 		$t,\neg r \imp \neg u$ && \\
$t, \neg u \imp r$ & 		$t, u \imp r$ && \\
$\neg r, \neg u \imp \neg t$ & 		$\neg r, u \imp \neg t$ &&
\etab
We first show that arguments $\Imp a_1$ and $\Imp a_2$ are in the grounded extension but $\Imp a_1,\Imp a_2 \imp p$ is not.

Consider first argument $\Imp a_1$. This argument has several defeaters. All of them combine the arguments for $p$ and $q$ to conclude either $r$ or $\neg r$ and then combine the resulting argument for either $r$ or $\neg r$ with the argument for $q$ into an argument for $\neg p$. This argument then is with $\Imp a_2$ combined into an argument for $\neg a_1$. These complex arguments all have a strict defeater, namely, $t$, since $t$ together with $r$ implies $s$ and $\neg s$ while $t$ together with $\neg r$ implies $u$ and $\neg u$. Note that $t$ is strictly preferred over all arguments it dlp-rebuts, sine $t$ is strict while all arguments it dlp-rebuts are defeasible.  Since $t$, being strict, has no defeaters by consistency of $\K$, there is a winning strategy in the grounded game for $\Imp a_1$, so $\Imp a_1$ is in the grounded extension.

The proof that  $\Imp a_2$ is in the grounded extension is entirely similar.

We next show that $\Imp a_1,\Imp a_2 \imp p$ is not in the grounded extension. Call this argument $A$. It is dlp-rebutted by argument $C$ of the form $\Imp q$. Since both $A$ and $C$ are defeasible, they defeat each other.  Next, observe that there is no strict defeater of $C$, since all dlp-rebuttals of $C$ need the argument for $p$ as a subargument, which is defeasible.
%since defeaters that use $\neg q$ need an argument for either $\neg a_1$ or $\neg a_2$ and so need a defeasible subargument.
So there is no winning strategy for $A$ in the grounded game, so $A$ is not in the grounded extension.
\end{example}
Note that this also yields a counterexample if the set of defeasible rules is totally ordered and arguments are compared with the weakest- or last-link argument ordering as in \citeN{cmo14}, since we can then give all four defeasible rules equal priority.

Note that various arguments in the example are contradictory in the sense of Definition~\ref{delp-arg}(2), so imposing DeLP's consistency condition on arguments (by requiring that the set of all conclusions of all their subarguments is indirectly consistent), excludes this counterexample for \DLP. However, other counterexamples for \DLP exist,  for instance, Example~\ref{counterex2} from Section~\ref{conssec}. If strict closure and indirect consistency are regarded as desirable, then this is worrying not just for \DLP but also for \ASPIC\ since with the original \ASPIC\ definition of rebuttal strict closure and indirect consistency can, as noted above, for the full case with preferences not be shown without allowing inconsistent arguments. We have now seen that replacing rebut with dlp-rebut does not change this, so as regards strict closure and indirect consistency the current state-of-the art for \ASPIC\ is still suboptimal.

%%%%%%%%%%%%%%%%%%%%%%%%%%%%%%%%%%%%%%%%%%%%%%%%%%%%%%%
\section{Differences in argument evaluation}\label{argeval}

While the \DLP definition of warrant is similar to grounded semantics, there are also differences, caused by the fact that the constraints on argument lines (Definition~\ref{accline}) do not coincide with the constraints on games in the game-theoretic proof theory for grounded semantics (Definition~\ref{dialogue}).
The choices made on the definition of argumentation lines by~\citeN{g+s04} establish constraints based on particular intuitions that were shown in examples.
However, we believe this definition of argumentation lines has some arguably counterintuitive consequences over the set of warranted literals.
We will discuss them below and then show that adopting grounded semantics avoids these counterintuitive consequences.

\subsection{Having to move a proper defeater after a blocking defeater}

Condition (4) of the definition of an acceptable argument line requires the move of a proper defeater if the previous argument was a blocking defeater, regardless whether the previous argument was a supporting or an interfering argument. This differs from the grounded game, in which the proponent must move strict defeaters while the opponent can move weak defeaters.

\begin{example}\label{ex1wrong}
Consider a dlp with $\SSet_f = \{p,s,u\}$, $\SSet_s = \emptyset$ and $\DD$ consisting of the rules of the following arguments:
% grs - \btab{llllll}
% grs - $A_1$: & $\langle \{p \Imp q\},q \rangle$ &&& $A_2$: & $\langle \{p \Imp q;~q \Imp r\},r \rangle$ \\
% grs - $B_1$: & $\langle \{s \Imp t\}, t \rangle$ &&& $B_2$: & $\langle \{s \Imp t;~t \Imp \neg r\}, \neg r \rangle$ \\
% grs - $C$: & $\langle \{u \Imp \neg t\}, \neg t \rangle$ &&& &
% grs - \etab
\btab{llllll}
$A_1$: & $\langle \{\drule{q}{p}\},q \rangle$ &&& $A_2$: & $\langle \{\drule{q}{p};~\drule{r}{q} \},r \rangle$ \\
$B_1$: & $\langle \{\drule{t}{s}\}, t \rangle$ &&& $B_2$: & $\langle \{\drule{t}{s};~\drule{\neg r}{t}\}, \neg r \rangle$ \\
$C$: & $\langle \{\drule{\neg t}{u}\}, \neg t \rangle$ &&& &
\etab
$A_2$ and  $B_2$ attack each other on $\neg r$ and $r$, while $C$ attacks $B_1$ and thus also $B_2$ on $t$. Assume an argument ordering that makes  $A_2$ and $B_2$ as well as $C$ and $B_1$ blocking defeaters of each other (for example, by not assigning any priority to the rules). Then $A_2$ is not warranted. Its dialectical tree consists of just one argument line, namely, $A_2,B_2$ and $A_2$ is marked $D$ in this tree. Note that the line cannot be extended with $C$, since $B_2$ is a blocking defeater of $A_2$ while $C$ is not a proper defeater of $B_2$.

Assume now an argument ordering in which $C$ and $B_1$ are still blocking defeaters of each other but in which $B_2$ is a proper defeater of $A_2$ (for instance, by giving all rules in $B_2$ priority over all rules in $A_2$). Then $A_2$ is warranted, since its dialectical tree again consists of just one argument line but now it is $A_2,B_2,C$ and $A_2$ is marked $U$ in this tree. Note that the line cannot be extended with $B_1$ since  $C$ is a blocking defeater of $B_2$ while $B_1$ is not a proper defeater of $C$.

So by strengthening $B$'s attack on $A$, $A$ turns from not warranted into warranted, which seems counterintuitive.
This example could be analyzed from the reverse perspective of $A$ being warranted and debilitating $B$ to a blocking defeater will lead to $A$ not being warranted.
This clash of intuitions becomes an interesting issue to study, and a possible avenue for doing so is to adopt the grounded game because supporting arguments have to be strict defeaters while interfering arguments can be weak defeaters.
According to the grounded game,  $A_2$ is not justified with the second argument ordering, since the line $A_2, B_2, C$ can be extended with $B_1$ after which $C$ is not allowed since it is not a strict defeater of $B_1$.
\end{example}

\subsection{The non-repetition rule}

Suppose that in line with our analysis of Example~\ref{ex1wrong} condition (3) of acceptable argument lines is changed to the effect that supporting arguments must be strict defeaters while interfering arguments can be weak defeaters. Then in Example~\ref{ex1wrong} argument $A_2$ is still warranted with the second argument ordering, since the line $A_2,B_2,C$ cannot be extended with $B_1$ for another reason: $B_1$ is a subargument of an argument already moved in the line, namely, $B_2$, so condition (2) of acceptable argument lines prevents extending the line with $B_1$.
So this condition  could also be changed by adopting the rule of the grounded game that only the proponent cannot repeat its arguments.
Moreover, this non-repetition rule might not be extended to proper subarguments of an already-moved argument, as shown by the following example.

\begin{example}\label{ex2wrong} Consider a dlp with $\SSet_f = \{f_1,f_2,f_3,f_4\}$, $\SSet_s = \emptyset$ and $\R_d$ consists of the rules of the following arguments. Assume also that arguments are ordered according to strict specificity relations between the conflicting rules.
% grs -  \btab{ll}
% grs - $A$: & $\langle \{f_1 \Imp p\}, p\rangle$ \\
% grs - $B$: & $\langle \{f_2 \Imp q;~q \Imp \neg p\}, \neg p\rangle$ \\
% grs - $C$: & $\langle \{f_3,f_4 \Imp r;~r \Imp s;~s,f_2\Imp \neg q\}, \neg q\rangle$ \\
% grs - $D$: & $\langle \{f_4 \Imp \neg r;~\neg r \Imp \neg s\}, \neg s\rangle$
% grs - \etab
\btab{ll}
$A$: & $\langle \{\drule{p}{ f_1} \}, p\rangle$ \\
$B$: & $\langle \{\drule{q}{ f_2};~\drule{\neg p}{ q}\}, \neg p\rangle$ \\
$C$: & $\langle \{\drule{r}{ f_3,f_4};~\drule{s}{r};~\drule{\neg q}{s,f_2}\}, \neg q\rangle$ \\
$D$: & $\langle \{\drule{\neg r}{ f_4};~\drule{\neg s}{\neg r}\}, \neg s\rangle$
\etab
Note that $A$ and $B$ weakly defeat each other, $C$ strictly defeats $B$ on its subargument for $q$, and $D$ strictly defeats $C$ by weakly defeating its subargument for $s$. Now there is a strict defeater of $D$, namely,
\btab{ll}
$E$: & $\langle \{\drule{r}{f_3,f_4}\}, r \rangle$
\etab
However, $E$ is a subargument of $C$, so if constraint (3) on argument lines is adopted in the grounded game, then the game loses completeness, since $A$ and $C$ are in the grounded extension. Note that the  non-repetition rule of the grounded game does not prevent the moving of $E$, since $E$ is not identical to $C$.
\end{example}

We next show that \DLP's non-repetition rule can in combination with the other \DLP constraints on argument lines make that the set of warranted arguments is not admissible in the sense of \citeN{dung95}.
\begin{example}\label{ex3wrong} Consider a dlp with $\SSet_f = \{f_1,f_2,f_3,f_4,f_5\}$, $\SSet_s = \emptyset$ and $\R_d$ consists of the rules of the following arguments. Assume also that arguments are ordered according to strict specificity relations between the conflicting rules.
% grs -  \btab{ll}
% grs - $A$: & $\langle \{f_1 \Imp p\}, p\rangle$ \\
% grs - $B$: & $\langle \{f_2 \Imp r;~r,f_5 \Imp s;~s \Imp t;~t,f_1 \Imp \neg p\}, \neg p\rangle$ \\
% grs - $C$: & $\langle \{f_3 \Imp u;~u \Imp \neg t\}, \neg t\rangle$ \\
% grs - $D$: & $\langle \{f_4 \Imp w;~w, f_3 \Imp \neg u\}, \neg u\rangle$ \\
% grs - $E$: & $\langle \{f_5 \Imp \neg s;~\neg s, f_4 \Imp \neg w\}, \neg w\rangle$
% grs - \etab
\btab{ll}
$A$: & $\langle \{\drule{p}{f_1}\}, p\rangle$ \\
$B$: & $\langle \{\drule{r}{f_2};~\drule{s}{r,f_5};~\drule{t}{s};\drule{\neg p}{~t,f_1}\}, \neg p\rangle$ \\
$C$: & $\langle \{\drule{u}{f_3};~\drule{\neg t}{u}\}, \neg t\rangle$ \\
$D$: & $\langle \{\drule{w}{f_4};~\drule{\neg u}{w, f_3}\}, \neg u\rangle$ \\
$E$: & $\langle \{\drule{\neg s}{f_5};~\drule{\neg w}{\neg s, f_4}\}, \neg w\rangle$
\etab
According to specificity, $B$ is a proper defeater of $A$, $C$ is a blocking defeater of $B$, $D$ is a proper defeater of $C$ and $E$ is a proper defeater of $D$. So $A,B,C,D,E$ is an acceptable argument line.

Here the argument line terminates, while there is a proper defeater of $E$, namely
\btab{ll}
$F$: & $\langle \{\drule{r}{f_2};~\drule{s}{r}\}, s\rangle$
\etab
But $F$ cannot be appended to the argument line since it is a subargument of $B$. So $A$ is warranted. Moreover, it is easy to see that $C$, which is a supporting argument for $A$, is not warranted, because of the argument line
%
% grs - \btab{ll}
% grs - $C$: & $\langle \{f_3 \Imp u;~u \Imp \neg t\}, \neg t\rangle$ \\
% grs - $D$: & $\langle \{f_4 \Imp w;~w, f_3 \Imp \neg u\}, \neg u\rangle$ \\
% grs - $E$: & $\langle \{f_5 \Imp \neg s;~\neg s, f_4 \Imp \neg w\}, \neg w\rangle$ \\
% grs - $F$: & $\langle \{f_2 \Imp r;~r \Imp s\}, s\rangle$
% grs - \etab
\btab{ll}
$C$: & $\langle \{\drule{u}{f_3};~\drule{\neg t}{u}\}, \neg t\rangle$ \\
$D$: & $\langle \{\drule{w}{f_4};~\drule{\neg u}{w, f_3}\}, \neg u\rangle$ \\
$E$: & $\langle \{\drule{\neg s}{f_5};~\drule{\neg w}{\neg s, f_4}\}, \neg w\rangle$\\
$F$: & $\langle \{\drule{r}{f_2};~\drule{s}{r}\}, s\rangle$
\etab
Note that here $F$ can be appended to $C,D,E$, since $A$ is not in this line.

In sum, we have that A is warranted even though there is no warranted supporting argument that defends $A$ against its defeater $B$. So we end up with a set of warranted arguments that is not admissible in the sense of \citeN{dung95}. If admissibility is accepted as a minimum rationality constraint on argument evaluation, then this is another reason to adopt grounded semantics for \DLP.
\end{example}
Garc\'{\i}a and Simari motivate their non-repetition rule with an example that has essentially the same structure as Example~\ref{excrossover}. In this example they want to prevent infinite argumentation lines. The grounded game indeed prevents this: the only possible game for $A_2$ is $P_1$: $A_2$, $O_1$: $B_2$ and the game terminates with a win by $O$ since $P$ cannot repeat its argument $A_2$.

Garc\'{\i}a and Simari motivate the subargument part of their non-repetition rule with a schematic example of the following form (leaving the rules implicit):

\begin{example}\label{circex1} Consider the following arguments, where each argument $X_1$ is a subargument of $X_2$:
\btab{ll}
$\langle \{A_1\}, p\rangle$ & $\langle \{A_2\}, \neg r\rangle$\\
$\langle \{B_1\}, q\rangle$ & $\langle \{B_2\}, \neg p\rangle$\\
$\langle \{C_1\}, s\rangle$ & $\langle \{C_2\}, \neg q\rangle$\\
$\langle \{D_1\}, \neg p\rangle$ & $\langle \{D_2\}, \neg s\rangle$
\etab
Assuming that the other constraints on argumentation lines are satisfied, they want to prevent the infinite line $A_2,B_2,C_2,D_2,A_1,B_2,\ldots$. In the grounded game this is indeed achieved, since the proponent is not allowed to repeat $C_2$ in attack on $B_2$.
\end{example}
We conclude that the non-repetition rule of the grounded game treats all of Garc\'{\i}a and Simari's examples in the way they want and avoids the arguably counterintuitive outcomes of \DLP in other examples.

\subsection{Concordance}

Constraint (2) on argument lines requires both the set of all supporting and the set of all interfering arguments in the line to be concordant, which means that the set of all rules of all these arguments must be consistent with the facts. If the dlp satisfies \citeANP{c+a07}'s~\citeyear{c+a07} rationality postulate of strict closure (saying that the set of all conclusions of all arguments in an extension must be indirectly consistent), then for the set of supporting arguments the requirement of concordance is a semantically redundant but computationally desirable addition to the grounded game: redundant since if an argument is warranted/justified, then all of proponent's arguments in a winning strategy for the argument will be in the grounded extension and will thus satisfy concordance; and desirable since it may prune the search space. However, otherwise concordance for supporting arguments can make a difference. In such cases there seems to be no clear reason why either adopting or not adopting concordance is better.

However, for the set of interfering arguments requiring concordance is arguably undesirable, as the following example shows.

\begin{example}\label{ex4wrong} Consider a \dlp\ with $\SSet_f = \{f_1,f_2,f_3,f_4\}$, $\SSet_s = \emptyset$ and $\R_d$ consists of the rules of the following arguments. Assume also that arguments are ordered according to strict specificity relations between the conflicting rules.
 %\btab{ll}
%$A$: & $\langle \{f_1 \Imp p\}, p\rangle$ \\
%$B$: & $\langle \{f_2 \Imp q;~q \Imp \neg p\}, \neg p\rangle$ \\
%$C$: & $\langle \{f_3 \Imp r;~r,f_2 \Imp \neg q\}, \neg q\rangle$ \\
%$D$: & $\langle \{f_4 \Imp \neg q;~\neg q \Imp \neg r\}, \neg r\rangle$
%\etab
\btab{ll}
$A$: & $\langle \{\drule{p}{f_1}\}, p\rangle$ \\
$B$: & $\langle \{\drule{q}{f_2};~\drule{\neg p}{q}\}, \neg p\rangle$ \\
$C$: & $\langle \{\drule{r}{f_3};~\drule{\neg q}{r,f_2}\}, \neg q\rangle$ \\
$D$: & $\langle \{\drule{\neg q}{f_4};~\drule{\neg r}{\neg q}\}, \neg r\rangle$
\etab
The dialectical tree for $A$ has just one line, viz.\ $A,B,C$. Note that $D$ cannot be appended to the line since $B$ and $D$ support contradictory (sub)conclusions $q$ and $\neg q$. So $A$ is warranted. Yet $C$, which supports $A$, is not warranted, since the line $C$ can be extended with $D$. Note also that $\{A\}$ is not an admissible set since it does not defend $A$ against $B$ while $\{A,C\}$ is not an admissible set, since it does not defend $C$ against $D$.
\end{example}
If admissibility is adopted as a minimum constraint on sets of warranted arguments, then this example shows that concordance for the set of interfering arguments is undesirable. In other words, arguing in favour of concordance for the set of interfering arguments requires arguing against admissibility as a minimum requirement on sets of warranted arguments. A less controversial solution is to adopt the game for grounded semantics, which, as shown above, is arguably also a good idea for other reasons.

% HENRY: We eliminated this paragraph because is not correct.
% -- grs It should be noted that \citeN{g+s14} and \citeN{g+s18} drop the concordance requirement for sets of interfering arguments, in agreement with the above analysis. \citeN{g+s04} argue in favour of concordance of sets of supporting arguments with an example in which an argument is in some argumentation line both a supporting and an interfering argument.

\begin{example}\label{circex2} Consider the following arguments, built from the rules in the previous example, where each argument $X_1$ is a subargument of $X_2$:
\btab{ll}
$\langle \{A_1\}, p\rangle$ & $\langle \{A_2\}, \neg r\rangle$\\
$\langle \{B_1\}, q\rangle$ & $\langle \{B_2\}, \neg p\rangle$\\
$\langle \{C_1\}, r\rangle$ & $\langle \{C_2\}, \neg q\rangle$
\etab
Assuming that the other constraints on argumentation lines are satisfied, they want to prevent the line $A_2,B_2,C_2,A_2$. In the grounded game this not prevented but this does not make $A_2$ warranted or the line infinite. The line can (as an argument game) only be continued with $P_3$: $B_2$, $O_3$: $C_2$ after which the game terminates with a loss by the proponent since he cannot repeat $A_2$. So the grounded game satisfies Garc\'{\i}a and Simari's intuitions about this example.
\end{example}

%
%%%%%%%%%

\subsection{Reformulating \DLP with grounded semantics}\label{groundedDeLP}

We have seen that \DLP's definition of warrant has some arguably counterintuitive outcomes, due to the particular constraints on acceptable argumentation lines.
We have also seen that adopting grounded semantics both avoids these outcomes and treats Garc\'{\i}a and Simari's motivating examples in the way they want.
Therefore and because of the similarities between \DLP's notion of warrant and the grounded argument game, it seems a good idea to develop a version of \DLP with its current account of warrant replaced with grounded semantics.
This can be done by replacing the current constraints on acceptable argument lines with the rules of the grounded game (where proponent and opponent arguments are, respectively, equated with supporting and interfering arguments), and by replacing proper and blocking defeat with strict defeat and defeat. Then given the equivalence noted at the end of Section~\ref{sec:DeLP}, either the grounded game or an accordingly modified definition of dialectical trees can be used to redefine warrant.
Thus adopting grounded semantics for \DLP would also establish clear links between \DLP and a large body of other research on formal argumentation.
Among other things, it would facilitate a study of the satisfaction of rationality postulates in \DLP.
Finally, a version of \DLP with grounded semantics would in fact adopt the semantics of~\citeN{s+l92}, which paper was the original source of inspiration for the development of \DLP and which, as noted above, proposes a notion of warrant that was by~\citeN{dung95} shown to be equivalent to grounded semantics.

The question arises whether the move to grounded semantics and leaving all other definitions as they are (let us call the resulting system \DLPgr) changes anything as regards strict closure and indirect consistency. As it turns out, the answer is no, since Example~\ref{counterex2} from Section~\ref{conssec} is also for \DLPgr a counterexample to satisfaction of strict closure and indirect consistency. The point is that neither $A_2$ nor $A_5$ has a defeater, so the grounded game for these arguments ends after the first move. So while the move to \DLPgr ensures admissibility of the set of warranted arguments, it does not guarantee strict closure and indirect consistency of the set of warranted conclusions. On the other hand, Proposition~\ref{direct-consistency} applies to \DLPgr, so the set of warranted conclusions is, as in original DeLP, guaranteed to be directly consistent.

\section{Correspondence results}\label{corrsec}

We next study correspondence results between aspects of \DLP, \DLPgr and \ASPIC. Of all these results, only Proposition~\ref{corrconsequence} will depend on \DLPgr; all other results hold for both \DLP and \DLPgr. Throughout this section we will implicitly assume corresponding defeasible logic programs and argumentation theories with the same language and the same sets of rules and facts. That is, we assume that $f \in \K$ just in case $f \in \SSet_f$, that $S \imp \varphi \in \R_s$ just in case $\srule{\varphi}{S} \in \SSet_s$ and that $S \Imp \varphi \in \R_s$ just in case $\drule{\varphi}{S} \in \DD$. Below we will leave the translation between the \ASPIC\ and \DLP notations implicit.

We first address the problem of finding a correspondence between \DLP arguments and \ASPIC\ arguments. To find such a correspondence, several  assumptions on \ASPIC\ argumentation theories are needed, since  \DLP has unlike \ASPIC\ minimality and consistency conditions on arguments. Accordingly, we define \emph{simplified} \ASPIC\ argumentation theories as those $AT$ in which all arguments $A$ are minimal in that if $\mathtt{Conc}(A) = p$ then there exists no argument $A'$ such that $\mathtt{Conc}(A') = p$  and $\mathtt{DefRules}(A') \subset \mathtt{DefRules}(A)$; and in which for all arguments $A$ the set $\mathtt{Conc}(\mathtt{Sub}(A)) \cup \R_s \cup \K$ is indirectly consistent.
\begin{lemma}\label{conclemma}
Let $AT = (AS,\K)$ be any \ASPIC\ argumentation theory. For any argument $A$ based on $AT$ it holds that $\mathtt{Conc}(\mathtt{Sub}(A))$ equals the set of all antecedents and consequents of any rule in $\mathtt{Rules}(A)$.
\end{lemma}
\begin{proof}
Immediate from the definition of an \ASPIC\ argument.
\end{proof}
\begin{proposition}\label{corrargs1}
Let $(\Pi,\Delta)$ be any defeasible logic program with a corresponding \ASPIC\ argumentation theory $(AS,\K)$. For any  \DLP argument $D = \langle R,p\rangle$ given $(\Pi,\Delta)$ there exists an \ASPIC\ argument $A$ for $p$ on the basis of $(AS,\K)$ with $\mathtt{DefRules}(A) = R$.
\end{proposition}
\begin{proof}
Let $D = \langle R,p\rangle$ be any  \DLP argument given $(\Pi,\Delta)$. Then there exists a defeasible derivation $Dd = L_1, \ldots, L_n = p$ of $p$ given $(\Pi,R)$. Assume without loss of generality that $Dd$ is minimal. We  prove by induction on the definition of defeasible derivations that for any element of $Dd$ there exists an \ASPIC\ argument $A$ on the basis of $(AS,\K)$ with $\mathtt{DefRules}(A) \subseteq R$.

There are two base cases. If $L_i$ is a fact, then $L_i \in \K$ so $L_i$ is an \ASPIC\ argument with $\mathtt{DefRules}(A) = \emptyset \subseteq R$. If $L_i$ is a presumption, then $\Imp L_i \in \R_d$ so $\Imp L_i$ is an \ASPIC\ argument with $\mathtt{DefRules}(A) = \{\Imp L_i \} \subseteq R$.

The induction hypothesis is that for all elements $L_i$ of $Dd$ such that there exists a rule $r$ in $\Pi \cup R$ with body $B_1, \ldots, B_m$ and head $L_i$ and such that all of  $B_1, \ldots, B_m$ precede $L_i$ in $Dd$ there exists an \ASPIC\ argument $C_j$ for any $B_j$ ($1 \leq j \leq m$) with $\mathtt{DefRules}(C_j) \subseteq R$.  For the induction step consider any such a  rule $r$ and let the \ASPIC\ arguments for  $B_1, \ldots, B_m$ be $C_1, \ldots, C_m$. Then if $r \in \Pi$ then $r \in \R_s$, so $C = C_1, \ldots, C_m \imp L_i$ is an \ASPIC\ argument with $\mathtt{DefRules}(C) = \mathtt{DefRules}(C_1) \cup \ldots \mathtt{DefRules}(C_m) \subseteq R$. Otherwise, $r \in R$ so $r \in \R_d$, so  $C_1, \ldots, C_m \Imp L_i$ is an \ASPIC\ argument with  $\mathtt{DefRules}(C) = \mathtt{DefRules}(C_1) \cup \ldots \mathtt{DefRules}(C_m) \cup \{r\} \subseteq R$.

Finally, to prove that  $\mathtt{DefRules}(A) = R$, assume for contradiction that there exists a rule $r \in R$ such that $r \not\in \mathtt{DefRules}(A)$. Then by Lemma~\ref{conclemma} it holds that $r$'s head is not in $\mathtt{Conc}(\mathtt{Sub}(A))$. But consider then the sequence $Dd'$ obtained by listing all elements of $\mathtt{Conc}(\mathtt{Sub}(A))$ in any order such that the bodies of any rule precede the rule's head. It is easy to verify that $Dd'$ is a defeasible derivation of $p$ given $(\Pi,\Delta)$. But $Dd' \subset Dd$, so $Dd$ is not minimal: contradiction.
\end{proof}

\begin{proposition}\label{corrargs2}
Let $(\Pi,\Delta)$ be any defeasible logic program with a corresponding \ASPIC\ argumentation theory $(AS,\K)$ that is simplified. For any \ASPIC\ argument $A$ for $p$ on the basis of $(AS,\K)$  there exists a \DLP argument $D = \langle \mathtt{DefRules}(A),p\rangle$ given $(\Pi,\Delta)$.
\end{proposition}
\begin{proof}
Suppose $A$ is an \ASPIC\ argument for $p$. There are two base cases. Assume first $A = p$. Then $p \in \SSet_f$ and $p$ is a defeasible derivation since $p$ is a fact. Moreover, since the \ASPIC\ $AT$ is simplified, $\{p\} \cup \R_s \cup \K$ is indirectly consistent, so (since $\Pi = \R_s \cup \K$), no pair of contradictory literals can be derived from $\Pi \cup \emptyset$. Finally, $\emptyset$ is obviously a minimal subset of $\Delta$ satisfying all this. So $D = \langle \emptyset,p\rangle$ is a \DLP argument for $p$ given $(\Pi,\Delta)$.

Assume next $A$ is of the form $\Imp p$. Then $p$ is a defeasible derivation since $\Imp p \in \Delta$ so $p$ is a presumption. Moreover, since $AT$ is simplified, there exists no strict \ASPIC\ argument for $p$ so $A$ is minimal in its set of defeasible rules. 
Then the proof that $D = \langle \{\Imp p\},p\rangle$ is a \DLP argument for $p$ given $(\Pi,\Delta)$ is similar as for facts.

The induction hypothesis is that for any \ASPIC\ argument $\{A_1,\ldots,A_m\} \imp\hspace*{-3pt}/\hspace*{-3pt}\Imp p$ there exist \DLP arguments for $\mathtt{Conc}(A_1),\ldots,\mathtt{Conc}(A_m)$ given $(\Pi,\Delta)$. 
For the induction step, consider any such \ASPIC\ argument. 
Then there exist defeasible derivations $Dd_j$ for all these conclusions $\mathtt{Conc}(A_j)$ ($1 \leq j \leq m$). 
Then clearly $Dd_1,\ldots, Dd_m,p$ is a defeasible derivation for $p$. 
Moreover, since the \ASPIC\ $AT$ is assumed to be simplified, $\mathtt{Conc}(\mathtt{Sub}(A))\cup \K$ is indirectly consistent and since the heads of all rules in $\mathtt{DefRules}(A)$ are in $\mathtt{Conc}(\mathtt{Sub}(A))$, no pair of contradictory literals can be derived from $\Pi \cup \mathtt{DefRules}(A)$ (recall that $\Pi = \R_s \cup \K)$. If there existed a \DLP argument $D = \langle R,p\rangle$ for $p$  such that $R \subset \mathtt{DefRules}(A)$, then by Proposition~\ref{corrargs1} there would exist an \ASPIC\ argument $A'$ for $p$ with $\mathtt{DefRules}(A') = R$. 
But then $A$ would not be minimal in its set of defeasible rules (note that $\mathtt{Conc}(\mathtt{Sub}(A'))  \cup \K$ is indirectly consistent since $D$ satisfies the consistency constraint on \DLP arguments).
So then $AT$ would not be simplified, which contradicts our assumption that it is simplified. 
Thus $D = \langle \mathtt{DefRules}(A),p\rangle$ is a \DLP argument for $p$ given $(\Pi,\Delta)$.
\end{proof}
There are counterexamples to Proposition~\ref{corrargs2} for non-simplified \ASPIC\ argumentation theories. Some counterexamples are due to \DLP's consistency constraint on arguments. Consider an \ASPIC\ $AT$ with $\K = \{p\}$ and $\R_d = \{ \Imp \neg p\}$. Then there are \ASPIC\ arguments $p$ for $p$ and $\Imp \neg p$ for $\neg p$ but the latter has no corresponding \DLP argument, since both $p$ and $\neg p$ can be defeasibly derived from $\Pi \cup  \{ \Imp \neg p\}$ (recall that $\K \in \Pi$). Other counterexamples are due to \DLP's minimality constraint on arguments. Consider an \ASPIC\ $AT$ with $\K = \{p\}$, $\R_s = \{p \imp q\}$ and $\R_d = \{ \Imp q\}$. Then the \ASPIC\ argument $\Imp p$ for $p$ has no corresponding \DLP argument, since $\langle \emptyset, q\rangle$ is a \DLP argument for $q$, so $\langle \{ \Imp q\}, q\rangle$ is not minimal.

It can be shown that the correspondence between \ASPIC\ and \DLP arguments is many-to-one.
\begin{proposition} Let $(\Pi,\Delta)$ be any defeasible logic program with a corresponding \ASPIC\ argumentation theory $(AS,\K)$ that is simplified.
\ben
\item For some \DLP arguments given $(\Pi,\Delta)$ there exist more than one corresponding \ASPIC\ argument on the basis of $(AS,\K)$.
\item For all \ASPIC\ arguments $(AS,\K)$ there exist a unique corresponding  \DLP argument given $(\Pi,\Delta)$.
\een
\end{proposition}
\begin{proof}
For (1) consider an example with $\SSet_f = \{p;r;\srule{q}{p};\srule{q}{r}\}$ and $\DD = \{\drule{s}{q}\}$. The \DLP argument $\{\drule{s}{q}\}$ for $s$ has two corresponding \ASPIC\ arguments $A_1 = [p \imp q] \Imp s$ and $A_2 = [p \imp r] \Imp s$.

(2) follows since each \ASPIC\ argument has a unique set of defeasible rules and uses all these rules, while \DLP arguments are defined by minimal sets of defeasible rules. Then the  construction in the proof of Proposition~\ref{corrargs2} clearly induces a unique \DLP argument.
\end{proof}

%Sometimes we denote the \DLP argument corresponding to \ASPIC\ argument $A$ with $A^D$ and an \ASPIC\ argument corresponding to \DLP argument $D$ with $D^A$. The following proposition is then given Proposition~\ref{corrargs} easily proven by induction on the definitions of $\mathtt{Sub}$ for both \DLP and \ASPIC\ arguments. \rood{HP: Alas, there are counterexamples: consider (in \ASPIC\ notation) $\R_s = \{p \imp r\}, \R_d = \{\Imp p; p \Imp q\}$. The \ASPIC\ argument $[\Imp p] \Imp q$ has just two subarguments, namely, itself and $\Imp p$. But the corresponding \DLP argument $\langle \{\Imp p; p \Imp q\},q\rangle$ has not just itself and $\langle \{\Imp p\},p\rangle$ as subarguments but also $\langle \{\Imp p\},r\rangle$. If there exists a defeating \DLP counterargument against the latter, then this also is a counterexample to Proposition~\ref{corrattacks1} for both rebutting and ua-rebutting attack. Maybe these counterexamples are excluded under assumptions such as the strict rules being closed under transposition. But if there are also preferences, then for defeat this also requires conditions on the argument ordering. So I think we should leave this for future research.}
%%
%\begin{proposition}\label{corrsub}
%{\bf *** DELETE SINCE INCORRECT ***} For any pair of corresponding \DLP argument $D$ and \ASPIC\ argument $A$ it holds that $\mathtt{Sub}(D) = S$ iff $\mathtt{Sub}(A) = \{D'^A \mid D' \in S\}$ and $\mathtt{Sub}(A) = S$ iff $\mathtt{Sub}(D) = \{A'^D \mid A' \in S\}$.
%\end{proposition}

We next prove correspondences with respect to the attack relations. To this end, we first define \DLP-counterparts of \ASPIC's rebutting and unrestricted rebutting attacks.

\begin{definition}[\textbf{a-rebutting and ua-rebutting attack}]
A \DLP argument $\langle A,p\rangle$ \emph{a-rebuts} a \DLP argument $\langle B,q\rangle$ on  $\langle B',q'\rangle$ if $\langle B',q'\rangle$ is a subargument of $\langle B,q\rangle$ and $p = -q'$ and $q'$ was derived in $\langle B',q'\rangle$ with a defeasible rule.

A \DLP argument $\langle A,p\rangle$ \emph{ua-rebuts} a \DLP argument $\langle B,q\rangle$ on  $\langle B',q'\rangle$ if $\langle B',q'\rangle$ is a subargument of $\langle B,q\rangle$ and $p = -q'$ and $B' \not= \emptyset$.
\end{definition}
In the  proofs of the following propositions we overload the symbol $\subseteq$ by writing for two \DLP arguments $D_1 = \langle S_1,p\rangle$ and $D_2 = \langle S_2,q\rangle$ that $D_1 \subseteq D_2$ to mean that $S_1 \subseteq S_2$. Likewise for other set-theoretic notations.
\begin{proposition}\label{corrattacks2}
For all \ASPIC\ arguments $A$ and $A'$ on the basis of a simplified argumentation theory it holds that if $A$ is (rebutted, u-rebutted, dlp-rebutted) by $A'$ then the \DLP argument $D$ corresponding to $A$ is (a-rebutted, ua-rebutted, rebutted) by the \DLP argument $D'$ corresponding to $A'$.
\end{proposition}
\begin{proof}
Suppose \ASPIC\ argument $A'$ for $p'$ rebuts, u-rebuts or dlp-rebuts  \ASPIC\ argument $A$ on $A''$ for $p$.  Consider the corresponding \DLP arguments $D'$ for $p'$ and $D''$ for $p$, which exist and are unique by Proposition~\ref{corrargs2}.  Note that $D'' \subseteq D$ by Proposition~\ref{corrargs2} since $\mathtt{DefRules}(A'') \subseteq \mathtt{DefRules}(A)$.

For rebut, $p = -p'$ while $p'$ was derived in $A''$ with $A''$ defeasible top rule. Since $A''$'s top rule is in $D''$,  $D'$ a-rebuts $D''$ since  $p$ and $p'$ are complementary literals. But then $D'$ a-rebuts $D$ since $D'' \subseteq D$.

For u-rebut, $p = -p'$ while $\mathtt{DefRules}(A'') \not= \emptyset$. Then by Proposition~\ref{corrargs2} it holds that $D'' \not= \emptyset$. Then $D'$ ua-rebuts $D''$ since $p$ and $p'$ are complementary literals. But then $D'$ ua-rebuts $D$ since $D'' \subseteq D$.

For dlp-rebut,  $\K \cup \{p,p'\}$ is indirectly inconsistent. Then in \DLP the set $\SSet \cup \{p,p'\}$ is contradictory.  Then $D'$ rebuts $D''$.  But then $D'$ rebuts $D$ since $D'' \subseteq D$.
\end{proof}
%The converse result can only be proven for a-rebut (in \DLP) and rebut (in \ASPIC). A counterexample for the other kinds of attack is (in \ASPIC\ notation) $\R_s = \{p \imp r\}, \R_d = \{\Imp p; p \Imp q, \Imp \neg r\}$. Consider the \DLP argument $D_1 = \langle \{\Imp p; p \Imp q\}, q\rangle$. It is rebutted by argument $D_2 = \langle \{\Imp \neg r\}, q\rangle$ but the \ASPIC\

\begin{proposition}\label{corrattacks1}
For all \DLP arguments $D$ and $D'$ based on any defeasible logic program it holds that if $D$ is (rebutted, a-rebutted, ua-rebutted) by $D'$ then all \ASPIC\ arguments $A$ corresponding to $D$ are (dlp-rebutted, rebutted, u-rebutted) by any \ASPIC\ argument $A'$ corresponding to $D'$.
\end{proposition}
\begin{proof}
Suppose $D'$ for $p'$ rebuts, a-rebuts or ua-rebuts $D$ on $D''$ for $p$. Consider any  \ASPIC\ argument $A$ corresponding to $D$ and $A'$ corresponding to $D'$ , which exist by Proposition~\ref{corrargs2}. By the same proposition, $\mathtt{DefRules}(A) = D$ and $\mathtt{DefRules}(A') = D'$.

For rebut,  $\SSet \cup \{p,p'\}$ is contradictory, i.e., there exists a defeasible derivation for two literals $h$ and $-h$ from this set. Note that such a defeasible derivation is strict since the set contains no defeasible rules. Then in \ASPIC\ the set $\K \cup \{p,p'\}$ is indirectly inconsistent.  By Lemma~\ref{conclemma} and the fact that $\mathtt{DefRules}(A) = D$, we have that $p \in \mathtt{Conc}(\mathtt{Sub}(A))$ and $p' \in \mathtt{Conc}(\mathtt{Sub}(A'))$. Then $A'$ dlp-rebuts $A$.

For a-rebut, $p$ and $p'$ are complementary literals while $p'$ was derived in $D''$ with a defeasible rule. Then $p = -p'$. By Lemma~\ref{conclemma} and the fact that $\mathtt{DefRules}(A) = D$, we have that $p \in \mathtt{Conc}(\mathtt{Sub}(A))$. Moreover, by construction of $A$, $p$ is derived in $A$ with a defeasible rule. Then $A'$ a-rebuts some subargument of $A$ so $A'$ a-rebuts $A$.

For ua-rebut, $p$ and $p'$ are complementary literals while $D'' \not= \emptyset$. Then $p = -p'$. By Lemma~\ref{conclemma} and the fact that $\mathtt{DefRules}(A) = D$, we have that $p \in \mathtt{Conc}(\mathtt{Sub}(A))$.  Then $A'$ a-rebuts some subargument of $A$ so $A'$ a-rebuts $A$.
\end{proof}
Propositions~\ref{corrargs1},~\ref{corrargs2},~\ref{corrattacks2} and~\ref{corrattacks1} together imply the following proposition (recall that \DLPgr is the variant of \DLP modified with grounded semantics).
\begin{proposition}\label{corrconsequence}
Let $(\Pi,\Delta)$ be any defeasible logic program with a corresponding \ASPIC\ argumentation theory $(AS,\K)$ that is simplified. Let the \DLPgr and \ASPIC\ orderings coincide in that for all \DLPgr arguments $D_1$ and $D_2$ it holds that $D_1 \prec D_2$ iff for all corresponding \ASPIC\ arguments $A_1$ and $A_2$ it holds that $A_1 \prec A_2$ and for all \ASPIC arguments $A_1$ and $A_2$ it holds that $A_1 \prec A_2$ iff for the corresponding \DLP arguments $D_1$ and $D_2$ it holds that $D_1 \prec D_2$. Let also \DLPgr attack be rebut (a-rebut, ua-rebut) iff \ASPIC\ attack is dlp-rebut (rebut, u-rebut). Then any \DLPgr argument $D$ is warranted iff all corresponding \ASPIC\ arguments $A$ are justified and any \ASPIC\ argument $A$ is justified iff  the corresponding \DLP argument $D$ is warranted.
\end{proposition}

\begin{proof}
(Sketch)

From left to right, consider any \DLP argument $D$ that is warranted. Then there exists a dialectical tree $T^D$ for $D$ with $D$ labelled $U$. Consider any corresponding dialectical tree $T^A$ of \ASPIC\ arguments obtained by replacing any supporting argument in $T^D$ by some corresponding \ASPIC\ argument and replacing any interfering argument in $T^D$ by all corresponding \ASPIC\ arguments. By Proposition~\ref{corrattacks1}  and our assumptions on the argument orderings, the defeat relations between \DLP arguments in $T^A$ are preserved as defeat relations between the corresponding \ASPIC\ arguments in $T^D$. It is left to prove that $T^A$ contains all defeaters of any of its supporting arguments. Assume for contradiction that some defeater $B$ of some supporting argument $A$ in $T^A$ is not in $T^A$. By Proposition~\ref{corrargs2} $B$ has a unique corresponding \DLP\ argument $B^d$. Then by Proposition~\ref{corrattacks2} and our assumptions on the argument orderings it holds that $B^d$ defeats the \DLP argument $A^d$ corresponding to $A$. But then $B^d$ is in $T^D$ so $B$ is in $T^A$. Contradiction.

From right to left the proof is similar but simpler since any \ASPIC\ argument has just one corresponding \DLP argument.
\end{proof}

\section{Conclusion}

In this paper we have made detailed comparisons between \DLP and \ASPIC\ as formalisms for rule-based argumentation. 
The comparisons especially focussed on inter-translatability, consistency and closure properties and intuitive adequacy. 
Computational and implementational issues were not discussed but from other sources, such as \citeN{m+p18} and \citeN{g+s18}, it is clear that these issues have received substantially more attention for \DLP than for \ASPIC.  
Our comparisons have hopefully contributed to a better understanding of  the two formalisms and their relations, similarities and differences.

To summarise our main findings,  we have first seen that \DLP's notion of rebutting attack and its consistency and minimality constraints on arguments are intuitively appealing and in some special cases more so than there \ASPIC\ counterparts. However, we have also seen that the \DLP definitions may not fully comply with Caminada and Amgoud’s rationality postulates of strict closure and indirect consistency in cases where \ASPIC\ satisfies these postulates.
In Subsection~\ref{Rationality.Postulates} we have included  a thorough discussion about these issues.

% HENRY: We referred to the subsection where the discussion is more complete.
% -- grs Arguing in defence of the \DLP definitions and constraints requires arguing that these postulates are not minimum requirements for conclusion sets of defeasible argumentation systems.
% -- grs Alternatively, it could be argued that while \DLP's definitions may not be optimal in general, they are so for several special cases.

Furthermore, we have argued that there are reasons to consider a variant of \DLP with grounded semantics, since its current notion of warrant arguably has counterintuitive consequences in some examples and in general leads to sets of warranted arguments that are not admissible. We have seen that both problems can be avoided by adopting the argument game for grounded semantics in \DLP.  A version of \DLP with grounded semantics would also be a return to the semantics of \citeN{s+l92}, which paper was the original source of inspiration for the development of \DLP. Arguing in defence of \DLP's current notion of warrant requires arguing, first,  that its treatment of Example~\ref{ex1wrong} is not counterintuitive and second, that admissibility is not a minimum requirement for sets of warranted arguments. Alternatively, if one agrees that Example~\ref{ex1wrong} is treated incorrectly by \DLP but not that admissibility must be satisfied by warrant,  then less substantial changes in the definition of acceptable argument lines might suffice.

Finally, we have under some minimality and consistency assumptions on \ASPIC\ arguments identified a one-to-many mapping between \DLP arguments and \ASPIC\ arguments in such a way that if \DLP is modified with grounded semantics, then the resulting \DLPgr's notion of warrant is equivalent to  \ASPIC's notion of justification. This result was proven for three alternative definitions of attack.

As for future research, since \DLPgr generates abstract argumentation frameworks, it can be investigated to which extent properties of \DLPgr depend on grounded semantics or are inherited by other semantics for abstract argumentation frameworks. Future research could also investigate whether incorporating \DLP's notion of rebutting attack and/or its consistency and minimality requirements on arguments in \ASPIC\ can be done in a way  that fully preserves the current results on how \ASPIC\ respects the various rationality postulates. Given the results in this paper, this would require further changes in the \ASPIC\ framework.

\bibliographystyle{acmtrans}
\bibliography{argument,delp}

\label{lastpage}
\end{document}